arXiv:1804.04988v1 [cs.CV] 13 Apr 2018

# Convolutional Neural Network for Brain MR Imaging Extraction Using Silver-Standards Masks

Oeslle Lucena[1,a], Roberto Souza[b,c], Letícia Rittner[a], Richard Frayne[b,c], Roberto Lotufo[a]

[a]*Medical Image Computing Laboratory (MICLab), Department of Computer Engineering and Industrial Automation, School of Electrical and Computer Engineering, University of Campinas, Campinas, São Paulo, Brazil*

[b]*Departments of Radiology and Clinical Neurosciences, Hotchkiss Brain Institute, University of Calgary, Calgary, Alberta, Canada*

[c]*Seaman Family Magnetic Resonance Research Centre, Foothills Medical Centre, Alberta Health Services, Calgary, Alberta, Canada*

## Abstract

Convolutional neural networks (CNN) for medical imaging are constrained by the number of annotated data required for the training stage. Usually, manual annotation is considered to be the "gold-standard'. However, medical imaging datasets with expert manual segmentation are scarce as this step is time-consuming and expensive. Moreover, the single-rater manual annotation is often used in data-driven approaches making the algorithm optimal for one guideline only. In this work, we propose a convolutional neural network (CNN) for brain magnetic resonance (MR) imaging extraction, fully trained with what we refer to as silver-standard masks. To the best of our knowledge, our approach is the first deep learning approach that is fully trained using silver-standards masks, having an optimal generalization. Furthermore, regarding the Dice coefficient, we outperform the current skull-stripping (SS) state-of-the-art method in three datasets. Our method consists of 1) a dataset with silver-standard masks as input, 2) a tri-planar method using parallel 2D U-Net-based CNNs, which we refer to as CONSNet, and 3) an auto-context implementation of CONSNet. CONSNet refers to our integrated approach, *i.e.*, training with silver-standard masks and us-

*Corresponding author

*Email address:* `oeslle@dca.fee.unicamp.br` (Oeslle Lucena)

*URL:* `miclab.fee.unicamp.br` (Oeslle Lucena)

ing a CNN architecture. Masks are generated by forming a consensus from a set of publicly available automatic segmentation methods using the simultaneous truth and performance level estimation (STAPLE) algorithm. The CNN architecture is robust, with dropout and batch normalization as regularizers. It was inspired by the 2D U-Net model published by the RECOD group. We conducted our analysis using three publicly available datasets: the *Calgary-Campinas-359* (*CC-359*), the LONI Probabilistic Brain Atlas (LPBA40), and the Open Access Series of Imaging Studies (OASIS). Five performance metrics were used in our experiments: Dice coefficient, sensitivity, specificity, Hausdorff distance, and symmetric surface-to-surface mean distance. Our results also demonstrate that we have comparable overall performance *versus* current methods, including recently proposed deep learning-based strategies that were trained using manually annotated masks. Our usage of silver-standard masks, however, reduced the cost of manual annotation, diminished inter-/intra-rater variability, and avoided CNN segmentation super-specialization towards one specific SS method found when mask generation is done through an agreement among many automatic methods. Moreover, usage of silver-standard masks enlarges the volume of input annotated data since we can generate annotation from unlabelled data. Also, our method has the advantage that, once trained, it takes only a few seconds to process a typical brain image volume using a modern graphics processing unit (GPU). In contrast, many of the other competitive consensus-building methods have processing times on the order of minutes.

---

## 1. Introduction

Magnetic resonance (MR) images are widely used in clinical medicine and medical research, especially in diagnosing and studying brain disorders. MR imaging exhibits excellent soft tissue contrast that is not usually found in other imaging modalities, such as x-ray or computed tomography. Therefore, brain MR scanning is broadly accepted as producing good visualization of brain structures [1]. Due to this fact, brain MR images are often used to diagnose a variety of brain diseases including acoustic neuroma, Alzheimer's disease and other neurodegernative conditions, cerebrovascular diseases (like brain aneurysm, arteriovenous malformations, stroke), and tumours [2].

Segmenting brain tissue from non-brain tissue (a process known as brain extraction or skull-stripping, SS) is a critical step in many MR brain image processing algorithms. After brain extraction, the analysis of brain regions are more easily and more accurately performed [2], thus, accurate brain segmentation is an essential, early processing step. In fact, it is typically the initial step in a wide range of brain MR imaging analyses, such as when segmenting tissue types [3], analyzing multiple sclerosis lesions [4], classifying Alzheimer's disease [5], assessing schizophrenia [6], monitoring the development or aging of the brain [7], and determining changes in volumes and shape of brain regions across many disorders [8, 9]. Normally, brain MR images present with unwanted non-brain tissues that challenging the SS method. Furthermore, the brain gyri and sulci (*i.e.*, the ridges and depression on the brain outer surface, respectively) can challenge current state-of-the-art SS methods [10]. New approaches are continually being proposed to overcome these and other limitations, suggesting that the study of SS techniques remains an active research field using either conventional methods [10–16] or, more recently, deep learning (DL)-based approaches [17, 18].

After the groundbreaking result of Krizhevsky et al. [19], DL, especially convolutional neural network (CNN) approaches, has become a commonly employed algorithmic approach to solve medical imaging problems [20]. DL methods are trained with labelled raw data to "automatically discover" the underlying mathematical representations needed for detection, classification and/or segmentation [21]. Commonly, training a CNN from scratch requires a large amount of correctly labeled data. Appropriate medical image datasets, however, are generally too small. Situations often arise because labeled data frequently require significant manual expert effort to complete a time-consuming and expensive task [22].

To reduce this cost, single rater, manual annotation is often used. However, manual annotation is known to vary, even among highly-trained experts [23, 24], and thus, be impacted by both inter- and intra-rater variability [25]. Additionally, the characteristics of MR data are complex and can be impacted by several other factors including contrast differences among scanners and changes in image spatial resolution, especially at border pixels lying between tissues. These and other issues lead to the presence of ambiguous voxels and label confusion during manual annotation by experts [25]. Finally, manual annotation guidelines are generally subjective. For example, there are more than a dozen protocols for hippocampal segmentation and different protocols have been show to provide up to 2.5-fold variation in volume estimates [26]. The best case scenario to mitigate variability in manual annotation is development of a consensus agreement approach that uses multiple expert annotations in order to generate a robust "gold-standard". However for tasks like SS, forming a consensus among multiple experts is impractical due to the linear increase in cost associated with each additional manual segmentation. Consensus approaches can also generate annotated data from an agreement of different annotation masks or from the output of different automatic methods [23, 25, 27]. These results are potentially very robust and are usually applied to improve automatic multi-atlas methods [28, 29].

### *1.1. Previous Work*

Traditional (*i.e.*, non DL-based) SS methods can be categorized into one of six main classes: 1) manual annotation, 2) intensity-based methods, 3) morphology-based methods, 4) deformable surface-based methods, 5) atlas-based methods, and 6) hybrid methods [2]. The gold-standard method, manual annotation, is usually done by an expert, often a radiologist or similarly highly trained user. Manual methods, unfortunately, are time consuming; experts often need to spend hours segmenting one brain image volume. Manual annotations are considered, thus, impractical for medical analysis in large-scale studies. the scond class, intensity-based methods [16], are fast, but lack robustness. They are very sensitive to local changes in image contrast, noise, and artifacts. Morphology-based methods [14], the third class, are also fast, but depend on parameters that are experimentally computed and related to size and shape of mathematical morphological operations [2]. Deformable surface model-based methods [11] are a class that use a balloon-like template that deforms to fit the brain based on gradient information. Although they can fit both the interior and exterior areas of the brain, these methods are very dependent on initialization of the balloon-like template. Atlas-based methods [13, 15] rely on image registration to an atlas template, making them time-consuming approaches and very dependent of the atlas geometry. Lastly, hybrid methods attempt to combine the best features of the previously described methods. They generally require longer processing times, but usually achieve optimal segmentation results [10, 12, 30].

DL-based segmentation is often performed using one of two predominant approaches: 1) voxel-wise networks using CNN architectures with fully connected layers that classify the central pixel in an image patch, and 2) fully convolutional networks (FCNs) [31] that segment the entire image at one feedforward step. Both methods have been implemented using both 2D and 3D architectures, but because 3D convolutions are computationally expensive, 2D convolutions are more commonly used. Although the first class of approaches have been frequently exploited due to their derivation from classification tasks, FCN perform better in retrieving spatial information from local and global features. They are also faster than voxel-wise networks [31, 32]. Moreover, FCNs can work with any sized input because their weights do not depend on the input size, a limitation of voxel-wise networks.

Recently, two DL-based SS methods have proposed in the literature. Both methods were validated in small publicly available datasets against manual annotation. Kleesiek et al. [17] proposed a voxel-wise 3D CNN for SS that we will refer to as 3D CNN. The 3D CNN is not too deep due to the cost of the 3D convolutions, limiting its learning capacity. Salehi et al. [18] presented the auto-net for brain extraction. They examined two approaches: 1) parallel voxel-wise networks and 2) parallel 2D FCN U-Net [32], each followed by an auto-context CNN classifier. Basically, this classifier takes the concatenation of the probability maps and feeds them as input data to another CNN following the auto-context algorithm presented in Tu and Bai [33]. Context information retrieval with CNNs have been explored in other medical segmentation tasks. Chen et al. [34] also presented an auto-context version of its VoxResNet approach, which is an architecture based on the ResNet [35] for brain structures segmentation. Kamnitsas et al. [36] used multi-scale 3D CNN with fully connected conditional random fields (CRFs) for brain lesion segmentation, but such CRFs are time consuming and have very limited neighborhood relations compared to the auto-context approach [33].

Consensus methods could be used to generate annotated data from different automatic methods. For example, Rex et al. [27] compared the results of their consensus methods combining results from automatic methods. They obtained a higher agreement rate than that of different segmentations done by two different experts. Recently, consensus masks have been used to generate what we refer to "silver-standard" masks. Souza et al. [37] evaluated the agreement between consensus predictions and manual labelled data in the *Calgary-Campinas-359* (*CC-359*) publicly dataset which silver-standards masks generated by the consensus algorithm simultaneous truth and performance level estimation (STAPLE) [23] were used. This work also suggested the usage of consensus masks for training convolutional neural networks (CNNs). Lucena et al. [38] have investigated and validated the usage of silver-standard masks in the CNN training stage for SS. These and other results allow us to generate labelled data without the need of manual annotation, augmenting our training input datasets, and improve generalization over training a CNN with only a single manual annotation.

### *1.2. Our Approach*

We present in this paper a CNN approach for brain MRI SS. Unlike other methods, ours is completely trained using silver-standards masks that are generated by forming the consensus between eight non DL-based automatic skull-stripping methods. Our method exists in two main implementations: 1) a tri-planar method using parallel 2D CNNs that we will refer to as CONSNet and 2) an auto-context variation of CONSNet that adapts an auto-context CNN in cascade with the tri-planar method. The term CONSNet refers to the complete approach (*i.e.*, training with silver-standard masks and using CNN architecture). Our analysis were conducted on three publicly available datasets: *Calgary-Campinas-359* (*CC-359*) [37], LONI Probabilistic Brain Atlas (LPBA40) [39], and the Open Access Series of Imaging Studies (OASIS) [40]. We validated our method against manual annotations available in twelve image sets in *CC-359* and all image sets in the LPBA40 and OASIS datasets. Five performance metrics were used: Dice coefficient, sensitivity, specificity, Hausdorff distance, and symmetric surface-to-surface mean distance. Furthermore, we compared the processing time of our method against publicly available state-of-the-art automatic SS methods and the consensus-based masks generated by the STAPLE.

The principle contributions in this work can be summarized as follows: Our proposed approach is:

- the first to be fully trained with consensus-derived silver-standard masks. This step reduces the cost requirement required for manual annotation. It also enlarges the training input data so that it is suitable for large-scale analysis from non-annotated data.

- the first truly big data-compatible method for SS. We used a publicly available dataset with 347 images for training and also for extracting patches for data augmentation. Over 100,000 patches were generated and used as input data.

- generalizable. It was trained using the *CC-359* dataset and was then validated using the LPBA40 and OASIS datasets. It was found to outperform most state-of-the-art methods including some DL-based approaches. Moreover, we applied a two-fold experiment in order to compare our performance on the same dataset against other published DL-based methods.

- completely open-source and will be made publicly available.

This report is organized as follows. In Section 2 we describe the three datasets and present all the materials and methods for the analysis pipelines. The results of our methods are compared against the publicly available automatic SS methods in Section 3 and discussed in Section 4. Lastly, Section 5 presents the conclusions of and future work derived from our studies.

## 2. Materials and Methods

### *2.1. Datasets*

Three publicly available datasets were used for this study. All datasets contained adult human MR brain images acquired using a T1-weighted volumetric method. Some variability in the image acquisition parameters was present within and between the three datasets. The *CC-359* was used for training and both the LPBA40 and OASIS dataset used for validation. A total of 476 subjects were included in the datasets (218 males, 258 females, $51.16 \pm 10.40$ years).

#### *2.1.1.* CC-359 *Dataset*

The *CC-359* [1] is a public dataset composed of image volumes in NIfTI format from 359 subjects (176 males, 183 females, $53.50 \pm 7.80$ years) acquired in the coronal image plane. Data were collected on scanners from three different vendors (General Electric (GE) Healthcare, Philips Medical Systems, and Siemens Healthineers) and at two magnetic field strengths (1.5 T and 3 T). Image volumes had a spatial resolution of $1.0 \times 1.0 \times 1.0$ mm$^3$ [37]. The *CC-359* dataset includes the original volumes, the consensus masks were generated for all subjects using the STAPLE algorithm (described in Section 2.3). In addition, manual annotations were performed on twelve randomly selected subjects (two for each vendor-field strength combination).

[1] http://miclab.fee.unicamp.br/tools

### 2.1.2. LPBA40 Dataset

The LPBA40 [2] dataset is composed of of 40 T1-weighted image volumes from healthy subjects (20 males, 20 females, 29.2 ± 6.3 years) and their corresponding manually labeled brain masks [39]. The scans were acquired on a GE 1.5T system with a spatial resolution of $0.86 \times 1.5 \times 0.86$ mm$^3$.

### 2.1.3. OASIS Dataset

We use the first two discs of the OASIS [3] dataset that consist of T1-weighted voumes from 77 subjects, with spatial resolution of $1.0 \times 1.0 \times 1.0$ mm$^3$ [40]. This dataset contains 55 females and 22 males (average age: 51.6 ± 24.7 years). Twenty subjects (30%) had Alzheimers disease. The masks in this dataset were segmented with an custom method based on registration to an atlas, and then revised by human experts [10]; they were not manually segmented. As a result, the quality of the masks provided with this dataset is relatively poor. Nonetheless, we choose this dataset so that we can compare our results against published results of Kleesiek et al. [17] and Salehi et al. [18].

## 2.2. Automatic Skull-stripping Methods

This work employed a series of eight state-of-the-art, non DL-based, automatic skull-stripping methods, as well as two DL-based methods. These methods were used to develop consensus-derived labelled data (non DL methods only, Section 2.3) and to analyze the performance of our proposed SS methods (Section 2.6). In alphabetical order, the eight non DL methods were: 1) Advanced Normalization Tools (ANTs) [15], 2) Brain Extraction based on non-local Segmentation Technique (BEAST) [13], 3) Brain Extraction Tool (BET) [11], 4) Brain Surface Extractor (BSE) [16], 5) Hybrid Watershed Approach (HWA) [12], 6) Marker Based Watershed Scalper (MBWSS) [14], 7) Optimized Brain Extraction (OPTIBET) [30], and 8) Robust Brain Extraction (ROBEX) [10]. An overview of non DL-based methods was provided in Section 1 and further details can be found in the cited references. In our analysis we used the default processing parameters detailed in the above citations.

The two DL-based methods are the 3D CNN [17] and auto-net [18]. The 3D CNN approach is a a voxel-wise network containing seven convolutional hidden layers, one convolutional soft-max output-layer, and the receptive field (*i.e.*, input for each predicted pixel) of this model is $53^3$ pixels [17]. For this method, we use the available source code and model. The results of a two-fold cross-validation experiment with the LPBA40 and OASIS datasets were used in our comparitive analysis.

[2] http://www.loni.usc.edu/atlases

[3] http://www.oasis-brains.org/app/template/Index.vm

A second DL approach was auto-net is a 2D FCN U-Net followed by an auto-context CNN classifier. Salehi et al. [18] describe their results with and without the auto-context CNN in a two-fold cross-validation experiment using the LPBA40 and OASIS datasets. Only these results were used as the authors did not provide source code.

### *2.3. STAPLE "Silver-standard" Consensus*

Consensus methods can be used to provide more reliable and accurate segmentation labeling in SS and other image processing tasks. These methods combine different segmentations and obtain more robust results [23, 25, 27, 41]. STAPLE is one such consensus-forming algorithm that uses an expectation-maximization algorithm to estimate the hidden (or true) segmentation as a probabilistic mask. The algorithm considers a collection of segmentations and computes a probabilistic estimate of the true segmentation and a measure of the performance level represented by each segmentation method [23]. This algorithm was used in this work to generate what we refer to as "silver-standard" segmentation masks. Theses brain masks are formed from STAPLE output (a probability mask) by applying a threshold of 0.5. In our study, STAPLE used masks resulting from the eight non DL-based automatic skull-stripping techniques previously described (Section 2.2).

We applied the STAPLE algorithm to the LBPA40 and OASIS dataset. It has already been applied to the *CC-359* dataset using the same protocols that we adopted. STAPLE was chosen in this work because the algorithm has been validated extensively through experiments [23, 42, 43]. Moreover, an open-source implementation of the algorithm was available (Insight Segmentation and Registration Toolkit (ITK) repository [44]). For the *CC-359* dataset, "silver-standard" masks for the *CC-347* subset were generated and used for CNN training. The consensus brain masks generated for the twelve subjects with "gold standard" manual annotation (*i.e.*, the *CC-12* subset) were compared against the manual annotations. For clarity, the "silver-standard" masks derived from the *CC-12* are referred to as STAPLE-12. For LPBA40 and OASIS datasets, the "silver-standard" masks were only compared against manually annotated data. These "silver-standards" are referred to as STAPLE-LPBA40 and STAPLE-OASIS.

### 2.4. Convolutional Neural Network Architecture and Implementation

Our CNN is a based on the 2D FCN U-Net architecture. The original U-Net architecture is a "U"-shaped network (contracting path, left side; expansive path, right side, Figure 1) composed of 23 convolutional layers [32]. Our implementation is a modification of the CNN architecture from RECOD Titans [45][4], which is composed of 20 convolutional layers. A similar network won the 2017 Melanoma Screening Challenge [5].

In our implementation, we removed the fully connected layers and used a fixed kernel size of $3 \times 3$. We adopted the RMSprop [46] with an initial learning rate of $1.0e^{-3}$ and an exponential decay of 0.995 after each epoch at the training stage. Additionally, the negative of the Dice coefficient (described in Section 2.6) was used as the loss function. Lastly, the implementation was built using Keras with Tensorflow [47] as a backend.

### 2.5. Brain Extraction Pipelines

We present two DL-based brain extraction methods: 1) the regular CONSNet that consists of a tri-planar method implemented with three parallel 2D CNN pipelines and 2) an auto-context version of CONSNet. Both methods are summarized pictorially in Figure 2. The pipeline presented in Figure 2a shows the steps required to perform the CONSNet prediction where its final output is calculated after applying a threshold to the average probability of the three CNN output probability maps. The pipeline presented in Figure 2b overviews the procedure used to obtain the auto-context CONSNet version. This version takes the output probability maps from the three CNNs as the input to a fourth CNN.

The key idea in our implementation is to perform 2D segmentation on a slice-by-slice basis for each image volume and repeat for each orthogonal orientation (*i.e.*, the axial, coronal, sagittal reformatted images). 3D segmentation is then done by reconstruction through the concatenation of the 2D predictions. Our approach is analogous to what is done manually by an expert when reviewing volumetric images. To manually segment an image volume, a human operator would start the task in one view (often sagittal), and toggle to the other orthogonal views (axial, coronal) to determine that the voxel/region in question is or is not brain tissue. The operator is undertaking a form of "label voting" by assessing from different perspectives. By adopting the tri-planar approach, we expect that the CONSNet prediction improves when compared to only processing one orientation.

[4] https://github.com/learningtitans/isbi2017-part1

[5] https://recodbr.wordpress.com/2017/03/16/recod-wins-international-competition-for-melanoma-classification/

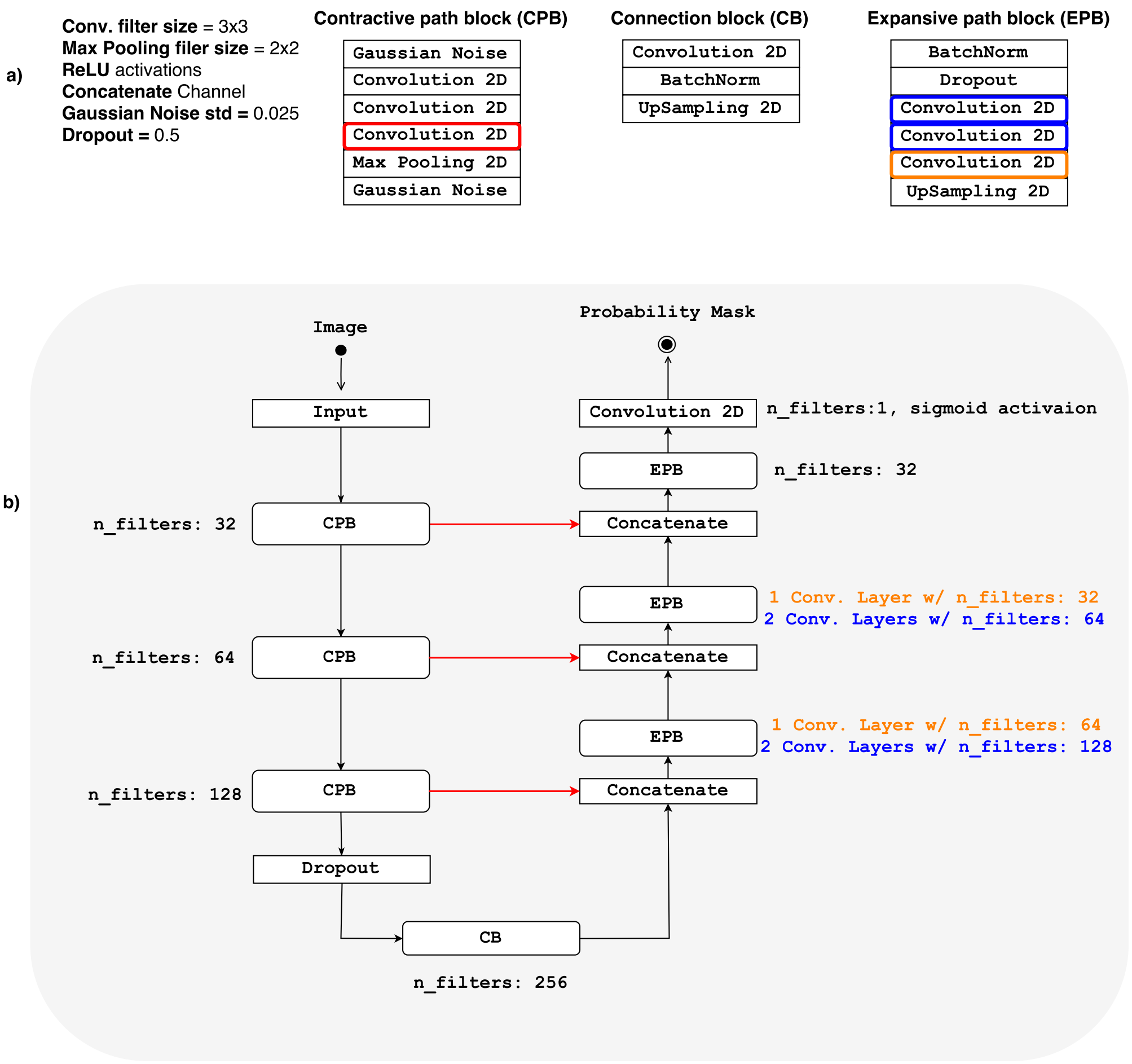

Figure 1: Modified RECOD U-Net architecture. The contracting path is on the left and the expansive path is on the right. Figure a) present configurations used in the architecture and composed blocks for the whole architecture draw in Figure b). b) draw of the whole architecture using the composed blocks, which are contractive path block (CPB), connection block (CB), and expansive path block (EPB). The concatenations followed by the red arrow are always done with the output of the third convolution layer (red in CPB block Figure a)) from the contracting path with the output of previous block at the expansive path. The third CPB block has a dropout instead of Gaussian noise layer and the two first EPB blocks have number of filters varying for each convolution. The legend in blue and orange in Figure b) correspond to the convolution layers with the same colors in the EPB block in a).

The auto-context CONSNet implmentation takes the probability maps generated by 2D parallel CNNs as input to a fourth CNN. The algorithm refines the output results in an iterative way, it integrates low-level and contextual information by fusing a large number of low-level appearance features with context and implicit shape information [34]. Originally Tu and Bai [33] used random forests classifiers to perform this step; here we take the advantage of the CNN architecture to extract the context information. In both cases, CONSNet and auto-context CONSNet, there are three major steps: 1) a pre-processing step, 2) a CNN segmentation step, and 3) a threshold and post-processing step. The details of each step are described in the following subsections.

#### *2.5.1. Pre-processing Step*

The *CC-347* subset, which is composed of 347 volumes and their respective "silver-standard" masks, was used to train our CNNs. Because that the range of grayscale intensities vary across the image volumes, we normalized the image volumes to be in the same intensity range (from 0 to 1000); a range chosen to ensure sufficient dynamic range and to minimize data storage limitations.

Secondly, because the volumes from different vendors have differing matrix dimensions, we varied the number of patches and their size for improved spatial content retrieval. In the end, three patches of size $128 \times 128$ were used. Patches were randomly extracted from each slice that contained brain voxels, and they were subsequently fed into the CNN at the training stage. This approach was applied in each axial, coronal and sagittal image plane for the CONSNet input training and after concatenation of the ouput probability maps for the auto-context CONSNet input. Note that we did not use patches surrounding a common voxel; rather, patches were randomly extracted across each image.

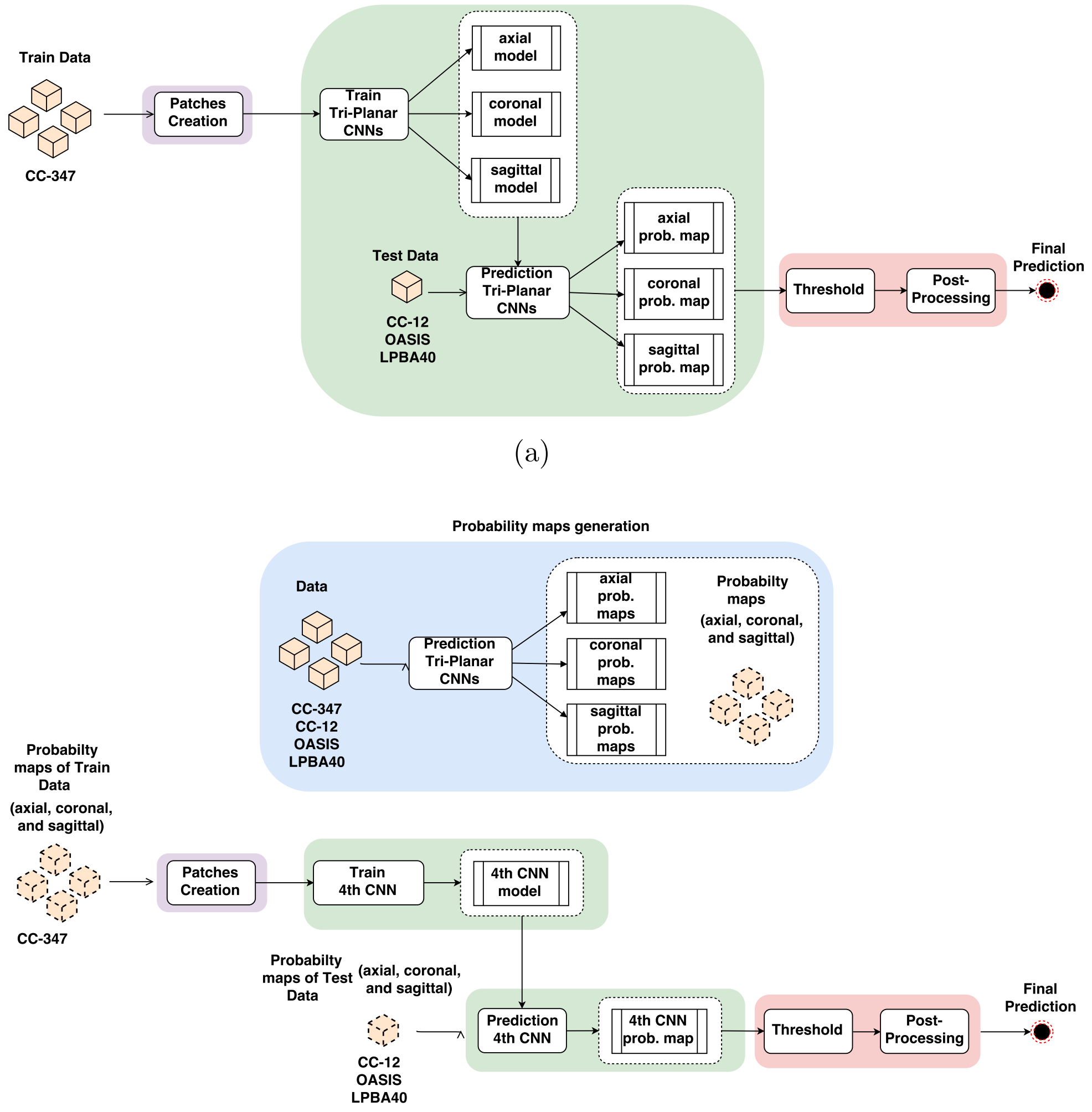

(a)

(b)

Figure 2: Deep learning brain segmentation pipelines. Both pipelines consist of three stages: pre-processing (purple), CNN segmentation (green), and threshold/post-processing (red). The CONSNet pipeline is shown in (a), and the auto-context CONSNet pipeline is shown in (b)

### *2.5.2. CNN Segmentation Step*

In the training step for CONSNet, the three 2D parallel CNNs, one for each axial, coronal, and sagittal image plane, were trained. The prediction step consisted of the predictions of all, one output image volume for each orthogonal model. For the auto-context CONSNet, the training step consisted of training the fourth CNN, but with three channels (one per orthogonal direction). This happens because we have to concatenate the output probability maps from the other CONSNet to produce its input. For this case, our CNN takes the input from the sagittal image plane. We chose that plane because it is the one often considered by the raters to perform most of the manual annotation. As mentioned before, they only toggle to the other orthogonal views to ensure whether the voxel/region in question is or is not brain tissue.

*2.5.3. Threshold and Post-processing Step*

In the first pipeline (Figure 2a), each of the CNN models considers different spatial information regarding the image plane that the CNN was trained. Therefore, to retrieve better spatial information from the predictions provided by the parallel CNNs, we took the average probability of the results, and then thresholded it. The threshold consisted of setting to one voxels that the average probability from the predictions of the three models are greater than 0.5. Otherwise, the voxels were set to 0.0. In the second pipeline (Figure 2b), there is only one prediction, thus, the threshold is applied directly in the CNN output. The voxels are thresholded at 0.5 only to produce a final prediction. The CONSNet and auto-context CONSNet final predictions are obtained after a post-processing step in which only the largest connected component was preserved. The other smaller components were removed using an area-open [48] filter implemented with the max-tree algorithm.

## *2.6. Evaluation Metrics and Statistical Analysis*

The metrics used to evaluate the segmentations were: Dice coefficient, sensitivity, specificity, Hausdorff distance, and mean symmetric surface-to-surface distance. The first three metrics are overlap metrics and the last two are surface distance metrics. Let $G$ be the ground truth image and $S$ the segmentation we wish to compare, the metrics can be defined by the following equations:

- Dice coefficient:

$$Dice(G, S) = \frac{2TP}{2TP + FP + FN}$$

- Sensitivity:

$$Sensitivity(G, S) = \frac{TP}{TP + FN}$$

- Specificity:

$$Specificity(G, S) = \frac{TN}{TN + FP}$$

- Hausdorff distance:

$$d_H(S, G) = \max\{\sup_{s \in S} \inf_{g \in G} d(s, g), \sup_{g \in G} \inf_{s \in S} d(s, g)\}$$

- Symmetric surface-to-surface mean distance:

$$d_S(S, G) = \frac{\sum_{s \in S} \min_{g \in G} d(s, g) + \sum_{g \in G} \min_{s \in S} d(g, s)}{|S| + |G|}$$

where $TP$, $FP$, $TN$, and $FN$ are number of true positives, false positives, true negatives, and false negatives, respectively, $d(\cdot, \cdot)$ is the euclidean distance, sup is supremum, and inf is the infimum.

Each adopted metric is responsible to evaluate different details regarding the prediction and ground through segmentations. Sensitivity measures how much brain tissue is included in the segmentation while specificity measures how much non-brain tissue is correctly segmented as non-brain. The Dice coefficient metric is a compromise between sensitivity and specificity, it evaluates the trade-of between the correct and false voxel predictions. The Hausdorff distance is an indicative of outliers, and the symmetric surface-to-surface mean distance has a similar interpretation to the Dice coefficient for distance between surfaces.

The statistical analysis to assess differences in the evaluation metrics was done using Wilcoxon signed-rank tests with Bonferroni correction. This test is a non-parametric statistical hypothesis test that does not assume normal distribution [49]. A $p$-value lower than 0.05 was used to assess statistical significance. For purposes of this statistical comparison, our auto-context CONSNet approach was selected as the reference method.

## 3. Experiments and Results

Our comparison analysis against the state-of-the-art methods were conducted into two experiments. In experiment 1) we trained the CONSNet and the auto-context CONSNet using the *CC-347* subset and we compared their results against the eight, non-deep learning, skull-stripping methods, the 3D CNN method, and the STAPLE as a method against the *CC-12* subset, OASIS and LPBA40 datasets. Remarks that for 3D CNN, we used the CNN model provided by its authors, which was trained in three public datasets (LPBA40, Internet Brain Segmentation Repository (IBSR), and OASIS). We have tried to trained the 3D CNN from scratch with our data, but the results were worse than with its provided model. Therefore, the authors model was applied to the test data from each adopted dataset in this work.

In experiment 2) we performed a 2-fold cross-validation in the LPBA40 and OASIS dataset to compare against the results of auto-net and the 3D CNN. We also validated against the *CC-12* subset and either OASIS or LPBA40, when one of them is used for the 2-fold cross-validation. In this experiment the size of patches extracted from each slice was different and set experimentally adding more information for the CNN according to the image-volumes shape from each dataset. In the end, we had ten patches of size $96 \times 96$ and size $144 \times 144$ for the LPBA40 and OASIS dataset, respectively. In this part, we intended to compare the robustness of our approach against the current deep-learning-based SS using gold standard masks. Also, we wanted to evaluate the model generalization by validating the 2-fold models against other datasets which data were not included in the training stage.

Tables 2, 4, and 3 summarize the overall analysis highlighting the two best results for each metric. Figures 3, 4, and 5 present *p-values* heat-maps for statistical significance results (*p-value* $< 0.05$ for each metric in the assessed test data. Also, box-plots related to the calculated metrics for each dataset are depicted in Figures 16, 17, 18, 19, and 20. The boxes in the plots are sorted in the ascending order with respect to the mean. Since BSE performed worst for the *CC-12* subset, we do not include its analysis for the LPBA40 and OASIS image volumes. Nonetheless, to generate the STAPLE as method, we included the BSE image volumes.

Table 1 has estimates prediction processing times for all assessed skull-stripping methods. Regard that the non-deep learning approaches are CPU-based, thus, for a fairly comparison, the processing times for our methods include both CPU and GPU evaluation. The value in front of the backslash represents the time computed on the CPU while the number after the backslash represents the time computed on the GPU. Remark that the processing time computed for the STAPLE does not include the time required to execute the automatic methods used as inputs; it only measured the processing time for the consensus. The same happened to the auto-context CONSNet processing time, which did not take account the time to generate the probability maps from the CONSNet prediction. Finally, These comparisons were done on workstation equipped with Intel Xeon® CPU E3-1220 v3 (3.10 GHz ×4) with 32 GB RAM and a NVIDIA GeForce Titan X GPU with 12 GB RAM.

Table 1: Approximate processing times for one image volume of each dataset (*CC-359*, OASIS, and LPAB40) for each method. ⋆ denotes results for the processing time for the STAPLE consensus as a method and auto-context CONSNet steps only (see text).

| **Processing time (seconds)** | | | |
|---|---|---|---|
| **Methods** | **Datasets** | | |
| | ***CC-359*** | **OASIS** | **LPBA40** |
| **ANTs** | 1378 | 1025 | 1135 |
| **BEAST** | 1128 | 944 | 905 |
| **BET** | 9 | 5 | 7 |
| **BSE** | 2 | 1 | 1 |
| **HWA** | 846 | 248 | 281 |
| **MBWSS** | 135 | 66 | 79 |
| **OPTIBET** | 773 | 579 | 679 |
| **ROBEX** | 60 | 53 | 57 |
| **3D CNN** | 196 | 121 | 123 |
| **STAPLE (12,OASIS,LPBA40)** | 160⋆ | 54⋆ | 36⋆ |
| **CONSNet** | 516/25 | 214/18 | 301/20 |
| **auto-context CONSNet** | 155⋆/11⋆ | 75⋆/8⋆ | 105⋆/10⋆ |

We also built heat maps to better visualize the segmented voxels as right or wrong of each method against the manual mask. First, the non-linear registration implemented in [15] was used to take all subjects to the same space. Then, we computed the FP and FN average error projection for all the skull stripping methods using the manually segmented subjects as reference and projected in the sagittal, coronal, and axial view.

The heat maps presented in Figures 10, 12, and 14 correspond to the FP projections, while the ones in Figures 11, 13, and 15 correspond to FN projections for all three datasets. The projections were normalized between 0 and 1. The upper extreme represents a high systematic number of FPs or FNs. Figures 6, 7, and 8 show representative 3D reconstructions of the different segmentation methods for one sample subject of the *CC-12*, OASIS, and LPBA40 datasets. The BSE reconstruction is not included due to the same reason mentioned before in the text.

Table 2: Overall analysis against manual segmentation results for the *CC-12* subset. The two best scores for each metric are emboldened.

| Methods | Metrics | | | | |
|---|---|---|---|---|---|
| | **Dice** (%) | **Sensitivity** (%) | **Specificity** (%) | **Hausdorff** (mm) | **Mean** (mm) |
| **ANTs** | $95.927 \pm 0.009$ | $94.51 \pm 0.016$ | $99.705 \pm 0.001$ | $\mathbf{8.905 \pm 1.393}$ | $0.057 \pm 0.015$ |
| **BEAST** | $95.766 \pm 0.012$ | $93.838 \pm 0.026$ | $\mathbf{99.757 \pm 0.001}$ | $9.907 \pm 1.41$ | $0.067 \pm 0.029$ |
| **BET** | $95.22 \pm 0.009$ | $98.261 \pm 0.016$ | $99.131 \pm 0.002$ | $12.169 \pm 2.766$ | $0.08 \pm 0.024$ |
| **BSE** | $90.488 \pm 0.070$ | $91.441 \pm 0.050$ | $98.648 \pm 0.020$ | $61.416 \pm 29.040$ | $1.562 \pm 3.179$ |
| **HWA** | $91.657 \pm 0.011$ | $\mathbf{99.93 \pm 0.001}$ | $97.83 \pm 0.008$ | $15.399 \pm 1.799$ | $0.179 \pm 0.038$ |
| **MBWSS** | $95.568 \pm 0.015$ | $92.784 \pm 0.027$ | $\mathbf{99.848 \pm 0.0004}$ | $28.228 \pm 5.446$ | $0.08 \pm 0.031$ |
| **OPTIBET** | $95.433 \pm 0.007$ | $96.133 \pm 0.01$ | $99.357 \pm 0.003$ | $10.304 \pm 1.998$ | $0.066 \pm 0.013$ |
| **ROBEX** | $95.611 \pm 0.007$ | $98.421 \pm 0.007$ | $99.13 \pm 0.003$ | $9.41 \pm 1.61$ | $0.063 \pm 0.015$ |
| **3D CNN** | $92.454 \pm 0.032$ | $88.77 \pm 0.059$ | $99.648 \pm 0.001$ | $21.244 \pm 14.921$ | $0.333 \pm 0.349$ |
| **STAPLE-12** | $96.797 \pm 0.007$ | $\mathbf{98.976 \pm 0.006}$ | $99.382 \pm 0.002$ | $\mathbf{8.327 \pm 1.665}$ | $0.038 \pm 0.007$ |
| **CONSNet** | $\mathbf{97.183 \pm 0.005}$ | $98.919 \pm 0.005$ | $99.465 \pm 0.002$ | $9.713 \pm 2.827$ | $\mathbf{0.037 \pm 0.007}$ |
| **auto-context CONSNet** | $\mathbf{97.191 \pm 0.005}$ | $98.944 \pm 0.005$ | $99.465 \pm 0.002$ | $9.137 \pm 2.374$ | $\mathbf{0.037 \pm 0.007}$ |

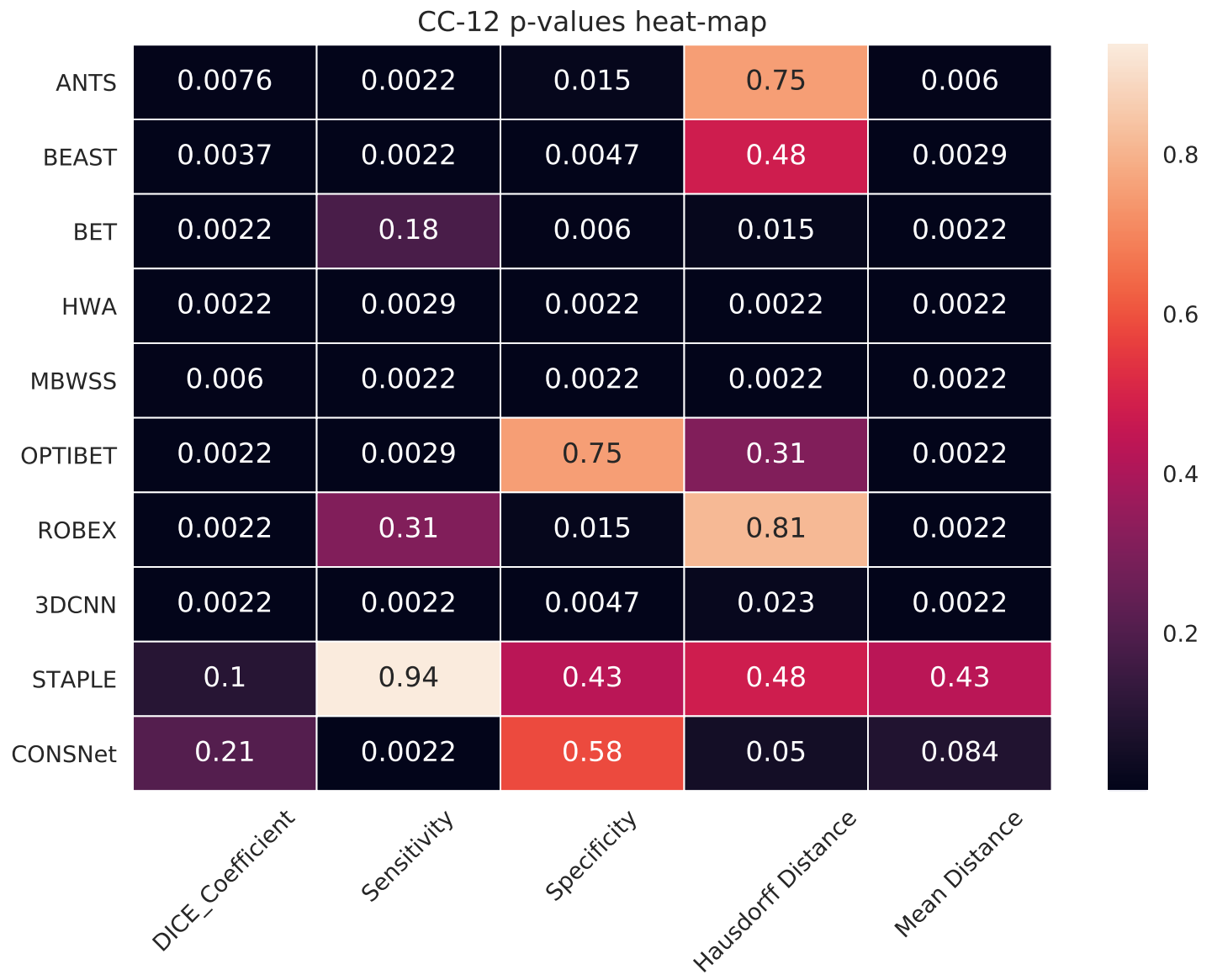

Figure 3: Heat-map of the *p-values* calculated using the auto-context CONSNet among all metrics assessed in the *CC-12* subset. Black cells highlight statistical significance (*p-values* $< 0.05$)

Table 3: Overall analysis against manual segmentation results for the LPBA40 dataset. The two best scores for each metric are emboldened.

| Methods | Metrics | | | | |
|---|---|---|---|---|---|
| | **Dice** (%) | **Sensitivity** (%) | **Specificity** (%) | **Hausdorff** (mm) | **Mean** (mm) |
| **ANTs** | $97.259 \pm 0.006$ | $\mathbf{98.981 \pm 0.004}$ | $99.179 \pm 0.002$ | $\mathbf{9.394 \pm 3.876}$ | $0.039 \pm 0.017$ |
| **BEAST** | $96.306 \pm 0.005$ | $94.06 \pm 0.012$ | $\mathbf{99.759 \pm 0.003}$ | $9.447 \pm 3.724$ | $0.058 \pm 0.016$ |
| **BET** | $96.625 \pm 0.007$ | $97.236 \pm 0.014$ | $99.276 \pm 0.002$ | $18.127 \pm 6.379$ | $0.079 \pm 0.065$ |
| **HWA** | $92.515 \pm 0.012$ | $\mathbf{99.898 \pm 0.001}$ | $97.092 \pm 0.006$ | $16.11 \pm 2.701$ | $0.206 \pm 0.055$ |
| **MBWSS** | $96.239 \pm 0.008$ | $94.406 \pm 0.013$ | $99.68 \pm 0.002$ | $23.661 \pm 6.283$ | $0.075 \pm 0.087$ |
| **OPTIBET** | $95.874 \pm 0.006$ | $93.349 \pm 0.011$ | $99.742 \pm 0.002$ | $12.536 \pm 2.838$ | $0.064 \pm 0.02$ |
| **ROBEX** | $96.773 \pm 0.002$ | $96.491 \pm 0.008$ | $99.469 \pm 0.002$ | $12.472 \pm 3.816$ | $0.05 \pm 0.006$ |
| **3D CNN** | $95.696 \pm 0.007$ | $92.614 \pm 0.015$ | $\mathbf{99.831 \pm 0.001}$ | $15.553 \pm 5.062$ | $0.07 \pm 0.021$ |
| **STAPLE-LPBA40** | $\mathbf{97.585 \pm 0.002}$ | $98.144 \pm 0.006$ | $99.457 \pm 0.002$ | $\mathbf{9.399 \pm 3.74}$ | $\mathbf{0.033 \pm 0.005}$ |
| **CONSNet** | $97.353 \pm 0.003$ | $97.257 \pm 0.007$ | $99.541 \pm 0.001$ | $12.35 \pm 3.721$ | $\mathbf{0.039 \pm 0.006}$ |
| **auto-context CONSNet** | $\mathbf{97.356 \pm 0.003}$ | $97.33 \pm 0.007$ | $99.528 \pm 0.001$ | $11.991 \pm 4.043$ | $0.039 \pm 0.007$ |

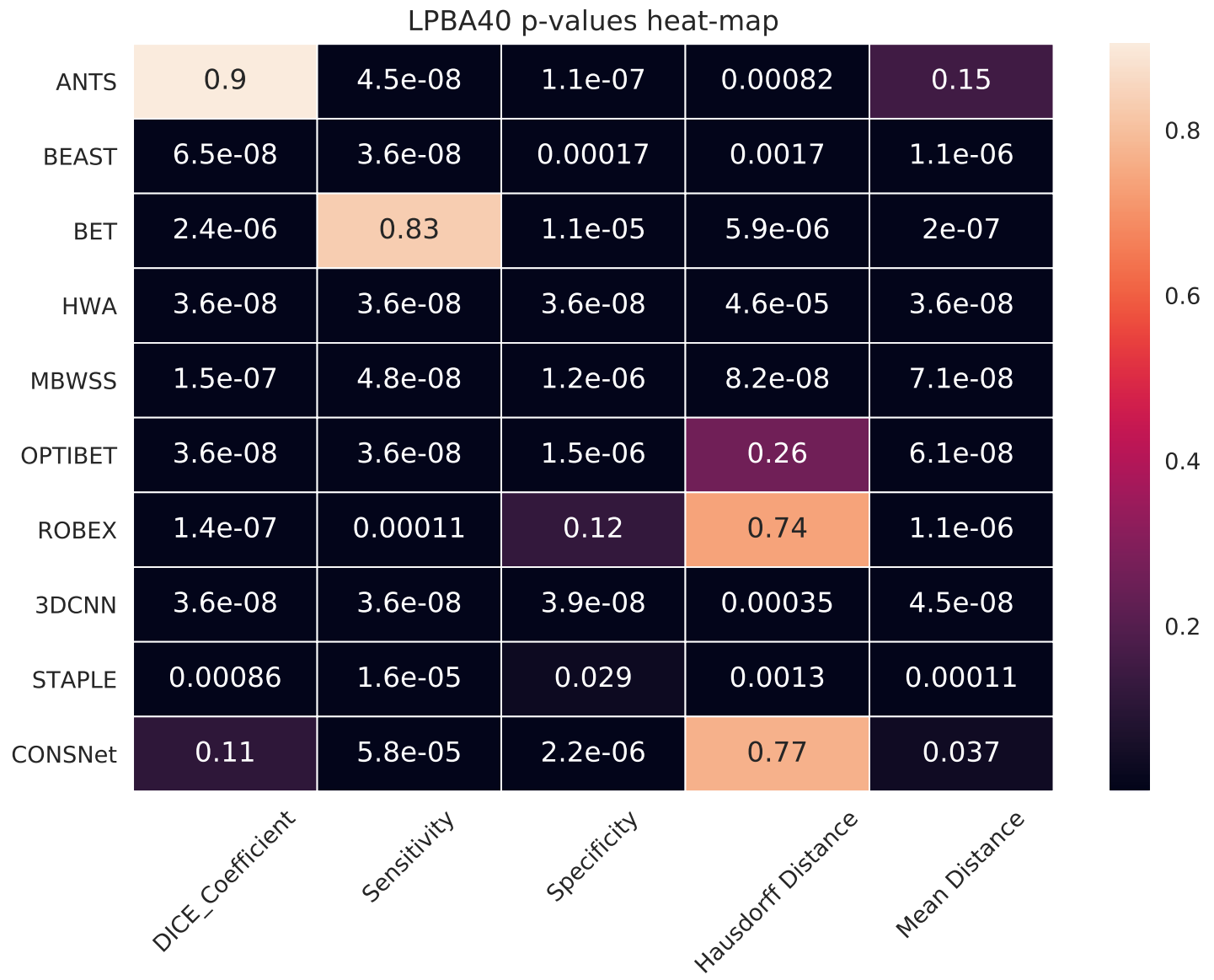

Figure 4: Heat-map of the *p-values* calculated using the auto-context CONSNet among all metrics assessed in the LPBA40 dataset. Black cells highlight statistical significance (*p-values* $< 0.05$)

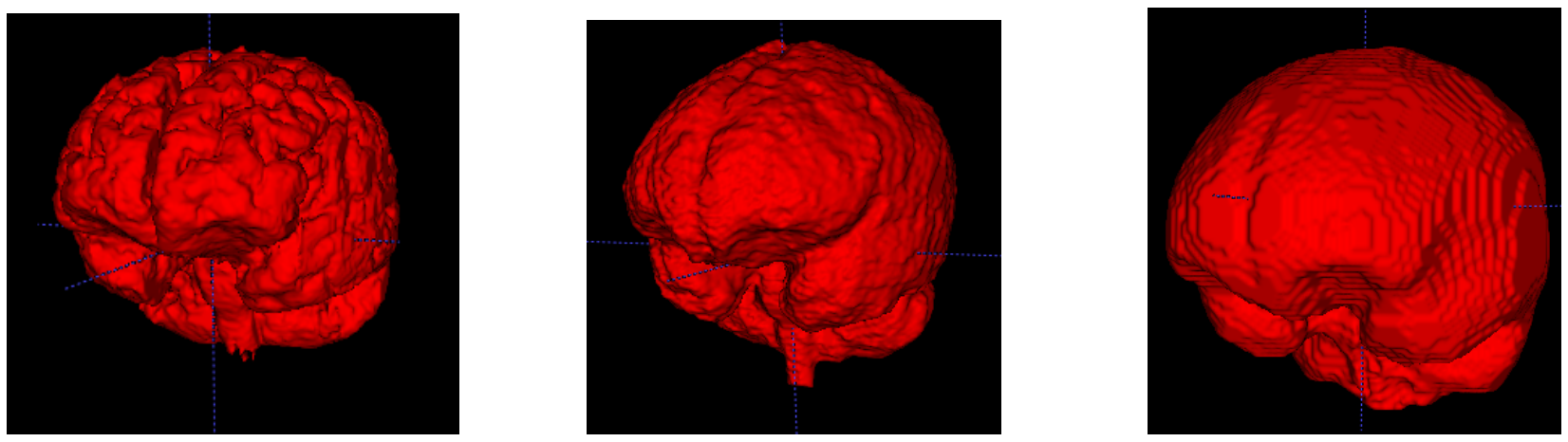

(a) Manual *CC-359* (b) Manual LPBA40 (c) Manual OASIS

Figure 9: Representative 3D reconstruction of the manual annotation for one subject of the *CC-359*, OASIS and LPBA40 datasets.

Table 4: Overall analysis against manual segmentation results for the *OASIS* dataset. The two best scores for each metric are emboldened.

| Methods | Metrics | | | | |
|---|---|---|---|---|---|
| | **Dice** (%) | **Sensitivity** (%) | **Specificity** (%) | **Hausdorff** (mm) | **Mean** (mm) |
| **ANTs** | $95.307 \pm 0.019$ | $94.391 \pm 0.036$ | $98.732 \pm 0.008$ | $9.898 \pm 4.35$ | $0.114 \pm 0.171$ |
| **BEAST** | $92.468 \pm 0.013$ | $86.763 \pm 0.025$ | $\mathbf{99.7 \pm 0.003}$ | $11.991 \pm 1.905$ | $0.167 \pm 0.039$ |
| **BET** | $93.503 \pm 0.027$ | $92.638 \pm 0.048$ | $98.101 \pm 0.013$ | $20.091 \pm 6.768$ | $0.227 \pm 0.242$ |
| **HWA** | $93.954 \pm 0.014$ | $\mathbf{98.36 \pm 0.015}$ | $96.125 \pm 0.016$ | $14.062 \pm 1.162$ | $0.149 \pm 0.055$ |
| **MBWSS** | $90.241 \pm 0.044$ | $84.094 \pm 0.079$ | $\mathbf{99.351 \pm 0.005}$ | $13.395 \pm 8.086$ | $0.249 \pm 0.297$ |
| **OPTIBET** | $94.456 \pm 0.011$ | $91.519 \pm 0.027$ | $99.222 \pm 0.005$ | $11.202 \pm 1.714$ | $0.11 \pm 0.031$ |
| **ROBEX** | $95.557 \pm 0.008$ | $93.954 \pm 0.022$ | $99.067 \pm 0.005$ | $\mathbf{9.442 \pm 1.813}$ | $\mathbf{0.083 \pm 0.025}$ |
| **3D CNN** | $95.237 \pm 0.009$ | $92.81 \pm 0.023$ | $99.277 \pm 0.004$ | $10.644 \pm 2.642$ | $0.095 \pm 0.031$ |
| **STAPLE-OASIS** | $\mathbf{96.096 \pm 0.007}$ | $\mathbf{95.188 \pm 0.02}$ | $98.983 \pm 0.006$ | $\mathbf{8.553 \pm 1.602}$ | $\mathbf{0.069 \pm 0.018}$ |
| **CONSNet** | $95.548 \pm 0.01$ | $93.98 \pm 0.028$ | $99.055 \pm 0.006$ | $10.228 \pm 3.932$ | $0.083 \pm 0.028$ |
| **auto-context CONSNet** | $\mathbf{95.602 \pm 0.01}$ | $94.021 \pm 0.028$ | $99.078 \pm 0.006$ | $9.614 \pm 3.658$ | $0.083 \pm 0.029$ |

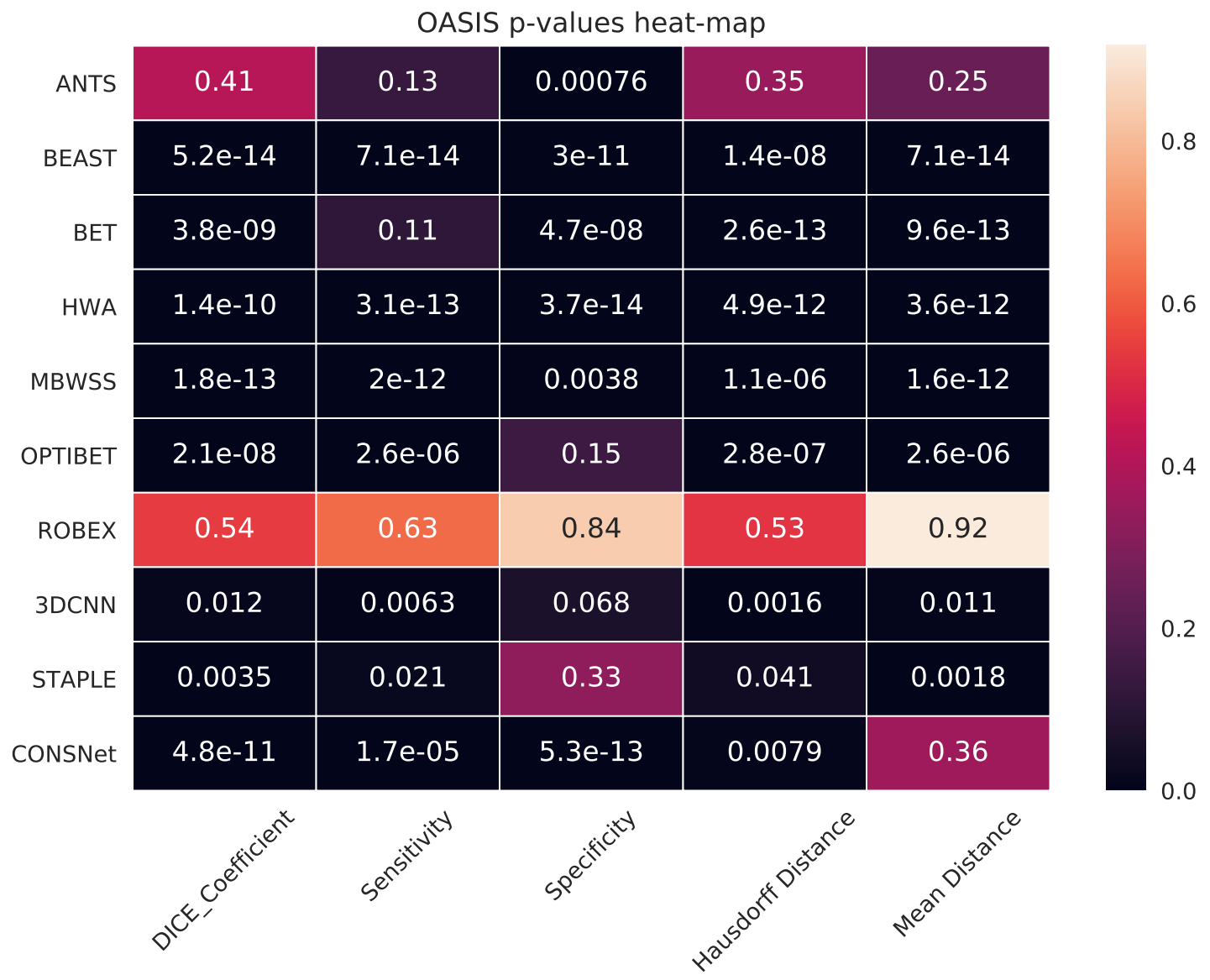

Figure 5: Heat-map of the *p-values* calculated using the auto-context CONSNet among all metrics assessed in the OASIS dataset. Black cells highlight statistical significance (*p-values* $< 0.05$)

# 4. Discussion

## *4.1. Experiment 1*

Regarding to the Dice coefficient which is the first metric taken by the raters to evaluate an optimal segmentation, our auto-context CONSNet out-performed the SS automatic algorithms presented in the literature (Tables 2, 3, and 4). That confirms our approach achieved optimal generalized performance with the LPBA40 and OASIS datasets and optimal robustness. Except in the OASIS dataset, which our CONSNet methd ranked second to ROBEX, generalized performance was optimal with the LPBA40 and OASIS datasets, and it had similar performance to its auto-context model. Remark that ROBEX segmentations are too smooth lacking brain tissue details (Figures 6, 8, and 7). Moreover, we outperformed the 3D CNN method in all three datasets which its model was trained using data from the LPBA40 and OASIS.

Table 5: 2-Fold for LPBA40 dataset. The best score for each metric is emboldened.

| Datasets | Methods | Metrics | | | | |
|---|---|---|---|---|---|---|
| | | **Dice** (%) | **Sensitivity** (%) | **Specificity** (%) | **Hausdorff** (mm) | **Mean** (mm) |
| **CC-12** | **CONSNet** | $89.63 \pm 0.076$ | $85.11 \pm 0.129$ | $99.6 \pm 0.002$ | $16.67 \pm 3.915$ | $0.24 \pm 0.217$ |
| **LPBA40** | **CONSNet** | $\mathbf{98.47 \pm 0.002}$ | $\mathbf{98.55 \pm 0.005}$ | $\mathbf{99.71 \pm 0.001}$ | $\mathbf{10.05 \pm 5.087}$ | $\mathbf{0.02 \pm 0.003}$ |
| | **Auto-net** [18] | $97.73 \pm 0.003$ | $98.31 \pm 0.006$ | $99.48 \pm 0.001$ | −− | −− |
| | **U-Net** [18] | $96.79 \pm 0.004$ | $97.22 \pm 0.016$ | $99.34 \pm 0.002$ | −− | −− |
| | **3D CNN** [17] | $96.96 \pm 0.01$ | $97.46 \pm 0.01$ | $99.41 \pm 0.003$ | −− | −− |
| **OASIS** | **CONSNet** | $92.55 \pm 0.03$ | $89.11 \pm 0.059$ | $98.86 \pm 0.007$ | $13.09 \pm 4.483$ | $0.15 \pm 0.075$ |

Table 6: 2-Fold for OASIS dataset. The best score for each metric is emboldened.

| Datasets | Methods | Metrics | | | | |
|---|---|---|---|---|---|---|
| | | **Dice** (%) | **Sensitivity** (%) | **Specificity** (%) | **Hausdorff** (mm) | **Mean** (mm) |
| **CC-12** | **CONSNet** | $92.22 \pm 0.022$ | $94.17 \pm 0.058$ | $98.92 \pm 0.004$ | $18.55 \pm 13.443$ | $0.17 \pm 0.087$ |
| **LPBA40** | **CONSNet** | $92.31 \pm 0.046$ | $90.78 \pm 0.08$ | $99.0 \pm 0.003$ | $17.8 \pm 8.306$ | $0.18 \pm 0.135$ |
| **OASIS** | **CONSNet** | $97.14 \pm 0.005$ | $97.45 \pm 0.013$ | $\mathbf{98.88 \pm 0.006}$ | $\mathbf{6.9 \pm 1.549}$ | $\mathbf{0.04 \pm 0.013}$ |
| | **Auto-net** [18] | $\mathbf{97.62 \pm 0.01}$ | $\mathbf{98.66 \pm 0.01}$ | $98.77 \pm 0.01$ | −− | −− |
| | **U-Net** [18] | $96.22 \pm 0.006$ | $97.29 \pm 0.01$ | $98.27 \pm 0.007$ | −− | −− |
| | **3D CNN** [17] | $95.02 \pm 0.01$ | $92.40 \pm 0.03$ | $99.28 \pm 0.004$ | −− | −− |

Our approaches had optimal results with respect to sensitivity *i.e.* keeping brain tissue in skull-stripping. All methods performed equivalently with the exception of HWA which usually has a high sensitivity but poor overlap and specificity due to a hard atlas registration. Regarding specificity, all methods performed optimally, except for the HWA (Figures 16, 17, 18, 19, and 20). Also, we had very few outliers measured by the Hausdorff and symmetric surface-to-surface mean distances (Tables 2, 3, and 4). Improvement for some metrics were statistical significant ($p$-$value < 0.05$) for almost all methods (Figures 3, 4, and 5).

From Table 1, it can be observed that the processing time of the most robust individual methods (ANTs, BEAST) was on the order of minutes for bigger image volumes like the ones from the *CC-12* subset. ROBEX which profits of parallelism, took around a minute for the same image volume. Our auto-context CONSNet prediction, under the same circumstances (CPU analysis), took a bit more than two minutes without considering the time to generate the probability maps from the parallel CNNs predictions. If we do consider that time, the auto-context CONSNet prediction will take $\approx 674$ seconds ($3 \times 173 + 155$, three times the time to generate the probability maps plus the time of the auto-context CONSNet prediction itself) which is still approximately twice faster than ANTs and BEAST. Furthermore, considering the prediction on a modern GPU process, ours required less than half a minute.

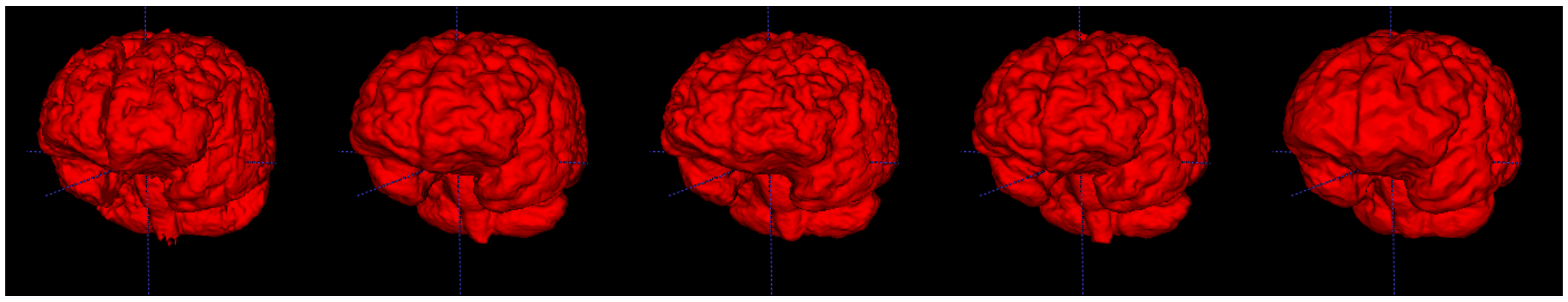

(a) Manual (b) STAPLE-12 $prob > 0.5$ (c) ANTs (d) BEAST (e) BET

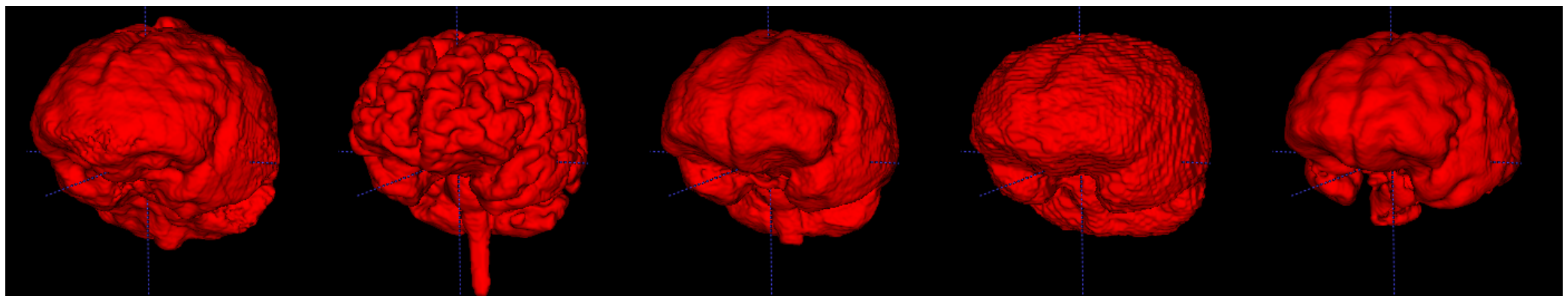

(f) HWA (g) MBWSS (h) OPTIBET (i) ROBEX (j) 3D CNN

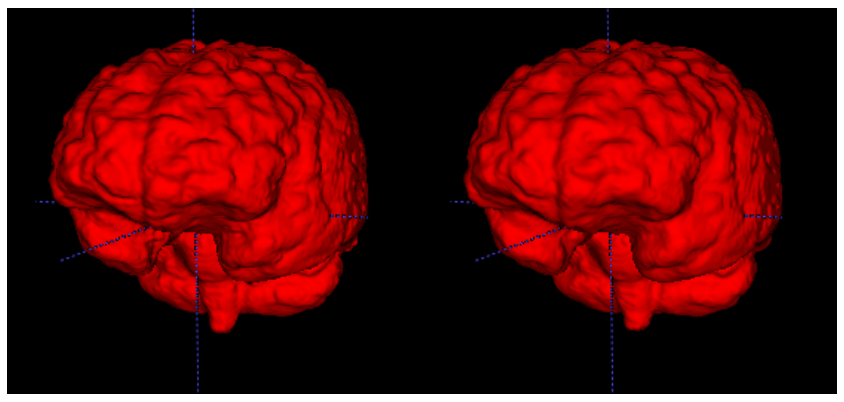

(k) CONSNet (l) auto-context CONSNet

Figure 6: Representative 3D reconstruction of the different segmentation methods for one subject of the *CC-12* subset.

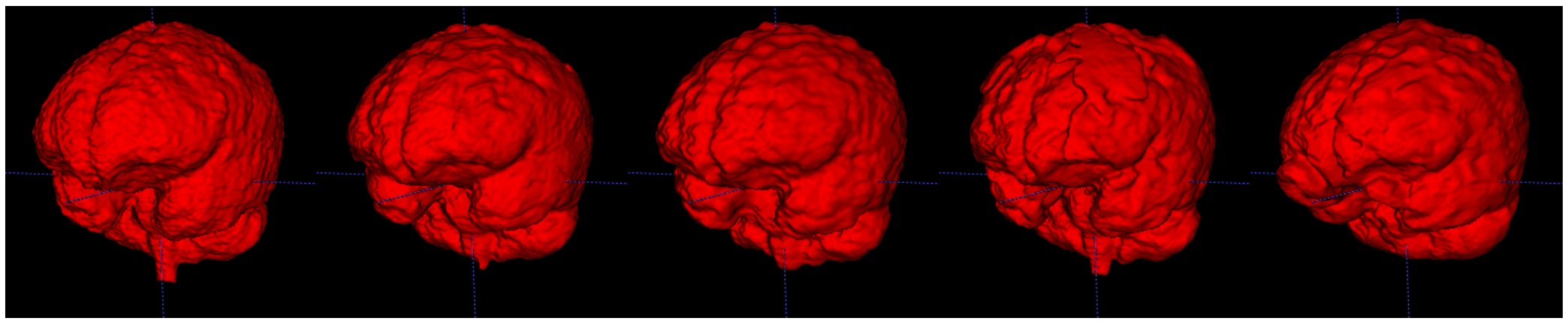

(a) Manual (b) STAPLE-LPBA40 $prob > 0.5$ (c) ANTs (d) BEAST (e) BET

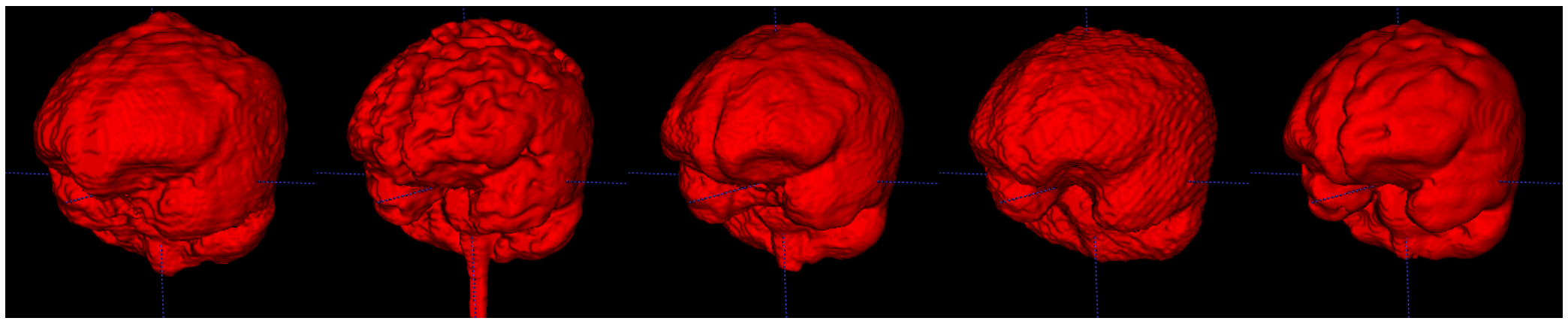

(f) HWA (g) MBWSS (h) OPTIBET (i) ROBEX (j) 3D CNN

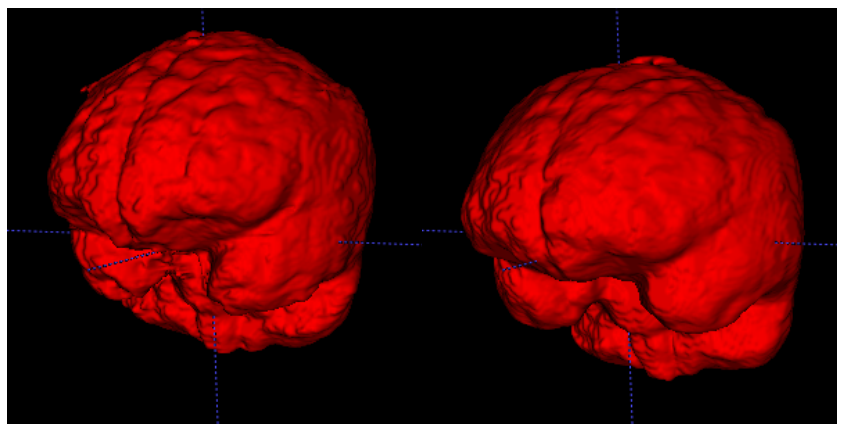

(k) CONSNet (l) auto-context CONSNet

Figure 7: Representative 3D reconstruction of the different segmentation methods for one subject of the LPBA40 dataset.

Figure 10 shows that even methods with high specificity, such as ANTs and BEAST were not able to properly segment the brain fissure between the left and right brain hemispheres in the *CC-12* subset, only MBWSS was capable of correctly segmenting the brain fissure. In the STAPLE consensus algorithm the influence of the MBWSS is one out of eight methods that did not correctly segmented the brain fissure affecting the silver-standard mask. Cpnsequently, our approach was more influenced from the other methods and also did not segmented the brain fissure correctly. Both OASIS and LPAB40 have annotated mask with smoothed brain fissures. From Figures 14 and 12 it is possible to see that have very few FP even among the brain border. 3D CNN method had the highest number of FP than the other methods.

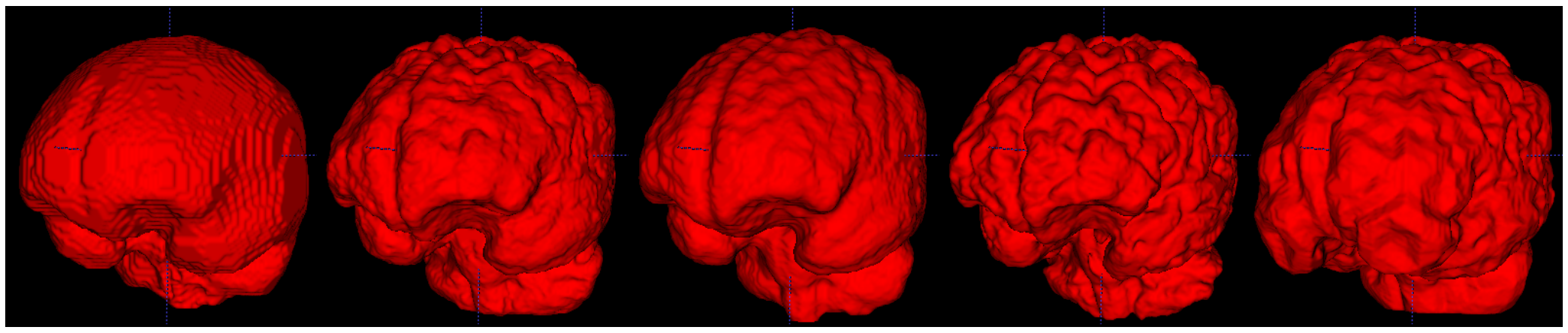

(a) Manual (b) STAPLE-OASIS $prob > 0.5$ (c) ANTs (d) BEAST (e) BET

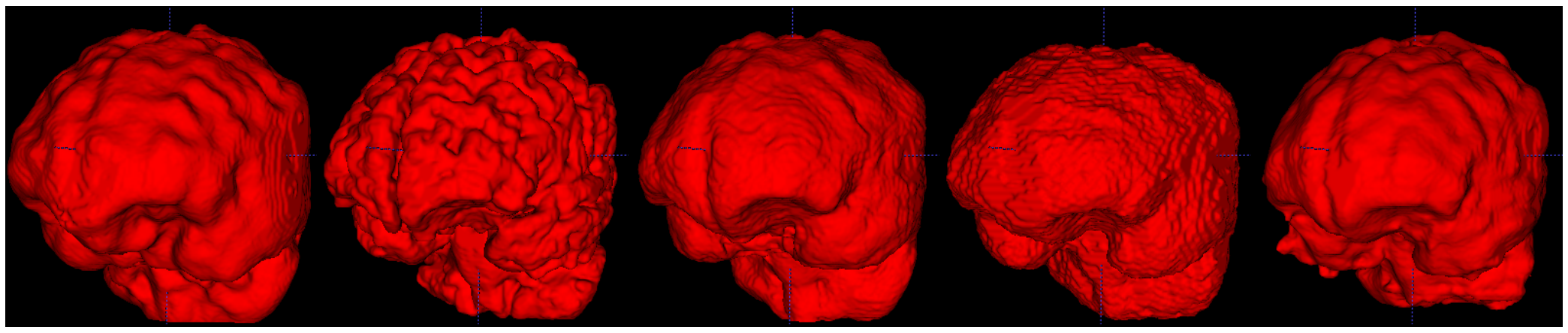

(f) HWA (g) MBWSS (h) OPTIBET (i) ROBEX (j) 3D CNN

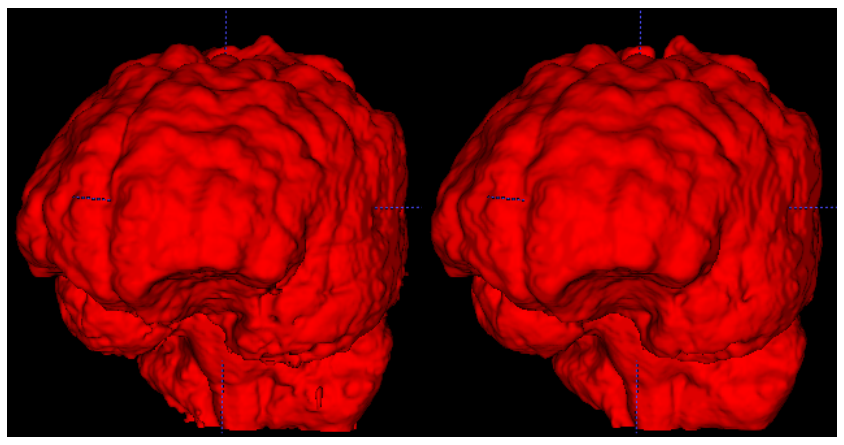

(k) CONSNet (l) auto-context CONSNet

Figure 8: Representative 3D reconstruction of the different segmentation methods for one subject of the OASIS dataset.

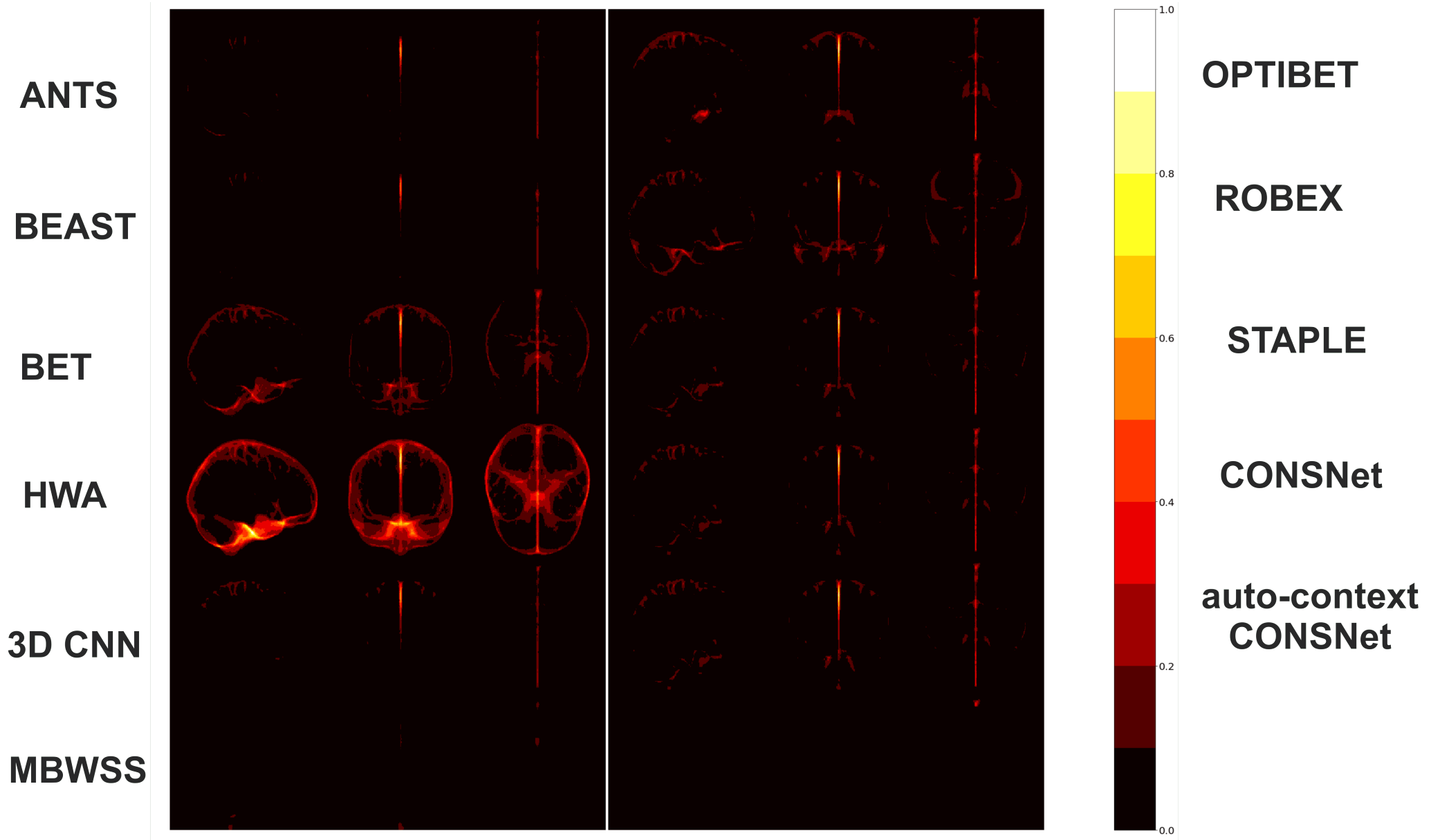

Figure 10: Sagittal, coronal and axial heat map projections for the *CC-12* subset of FP using the manual segmentations as reference.

Figures 11, 15, and 13 show that we had few FN in all datasets, mostly in the OASIS dataset. Furthermore, STAPLE had similar results to our method relatively to the heat maps analysis, except that STAPLE outperfomed ours in the OASIS and LPBA40 datasets.

The guidelines for manual annotation varies from expert raters, influencing the performance evaluation of the predictions for different datasets [13]. The *CC-12* subset had a manual annotation which was possible to correctly segment the gyri and sulci regions and the brain fissure, while the LPBA40 dataset had a smooth delineation, and the OASIS dataset has annotated data first automatically segmented, and then manually revised (Figure 9). Our silver-standard masks has the agreement among different automatic guidelines, overcoming the possible super-specialization in the CNN training towards one guideline resulting in an optimal performance with LPBA40 and OASIS datasets.

STAPLE as a method outperformed our method in almost all of the computed metrics in the OAISS and LPBA40 dataset (Tables 3 and 4) with statistical difference (Figures 4, and 5). Nonetheless, perform STAPLE as a method is very expensive and time-consuming because for every new im-age volume the automatic methods need to be ran again to generate a new STAPLE brain mask.

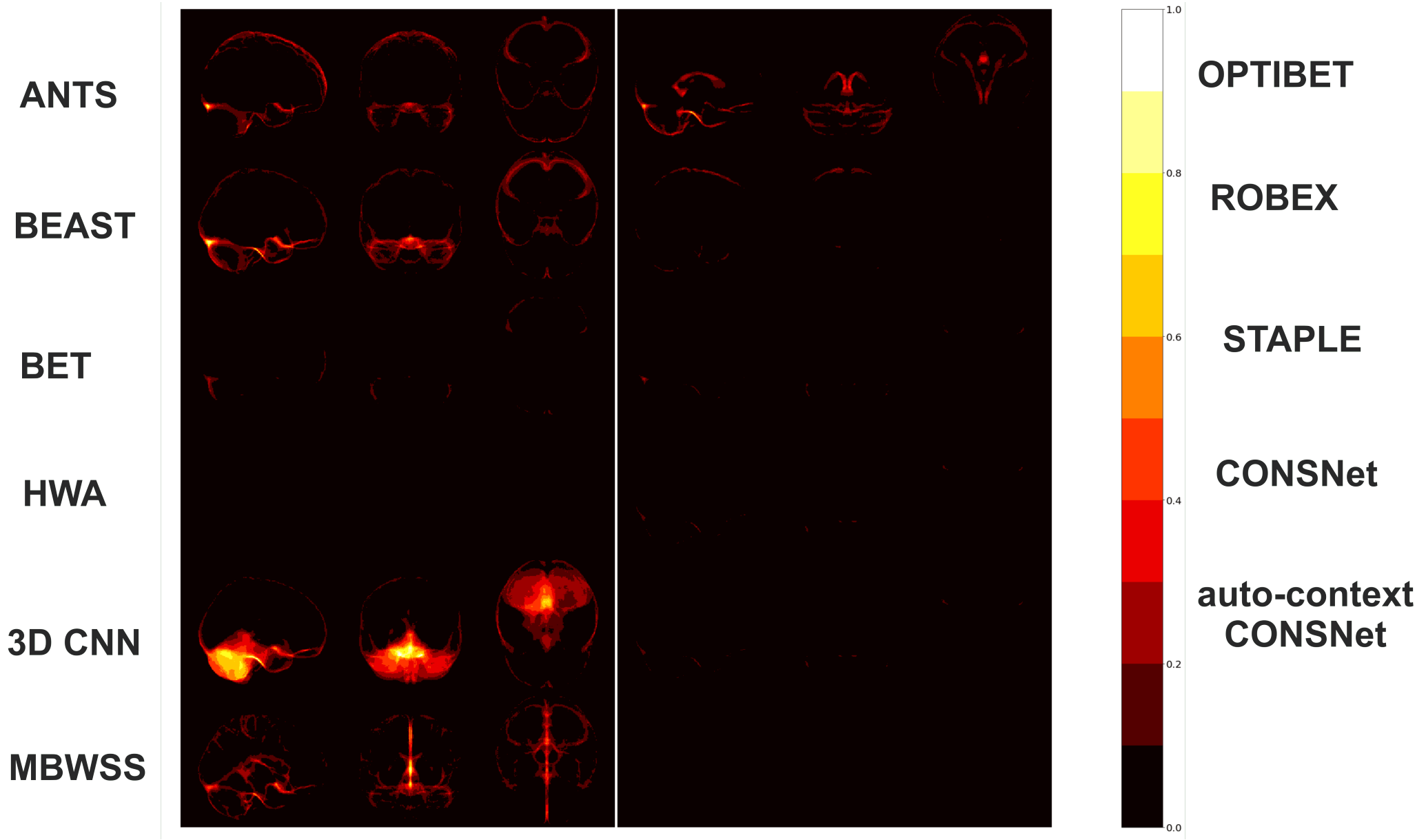

Figure 11: Sagittal, coronal and axial heat map projections for the *CC-12* subset of FN using the manual segmentations as reference.

### *4.2. Experiment 2*

The experiment 2 consisted of a 2-fold cross validation in the LPBA40 and OASIS dataset. For this case, we adopted the CONSNet only since the approach had similar results to the auto-context CONSNet with less cost in the experiment 1. Our CONSNet had comparable performance to the auto-net and outperformed the U-Net and the 3D CNN (Tables 5 and 6).

We outperform all DL-based methods in the LPBA40 dataset in all metrics. In the OASIS dataset, we ranked second regarding Dice coefficient and sensitivity and we ranked first regarding specificity in the OASIS dataset. The U-Net pipeline from Salehi et al. [18] work does not have an auto-context CNN. Therefore, a fairly comparison between CNN architectures is against U-Net which we outperformed with margin of 1% in OASIS 1.7% in LPBA40 datasets. Our CNN architecture is very robust, having Gaussian noise, dropout, and batch normalization as regularizers which lead to a substantial improvement against U-Net [18] and have comparable results to auto-net. It expected in this case that auto-context CONSNet achieve better results but with a cost of another CNN.

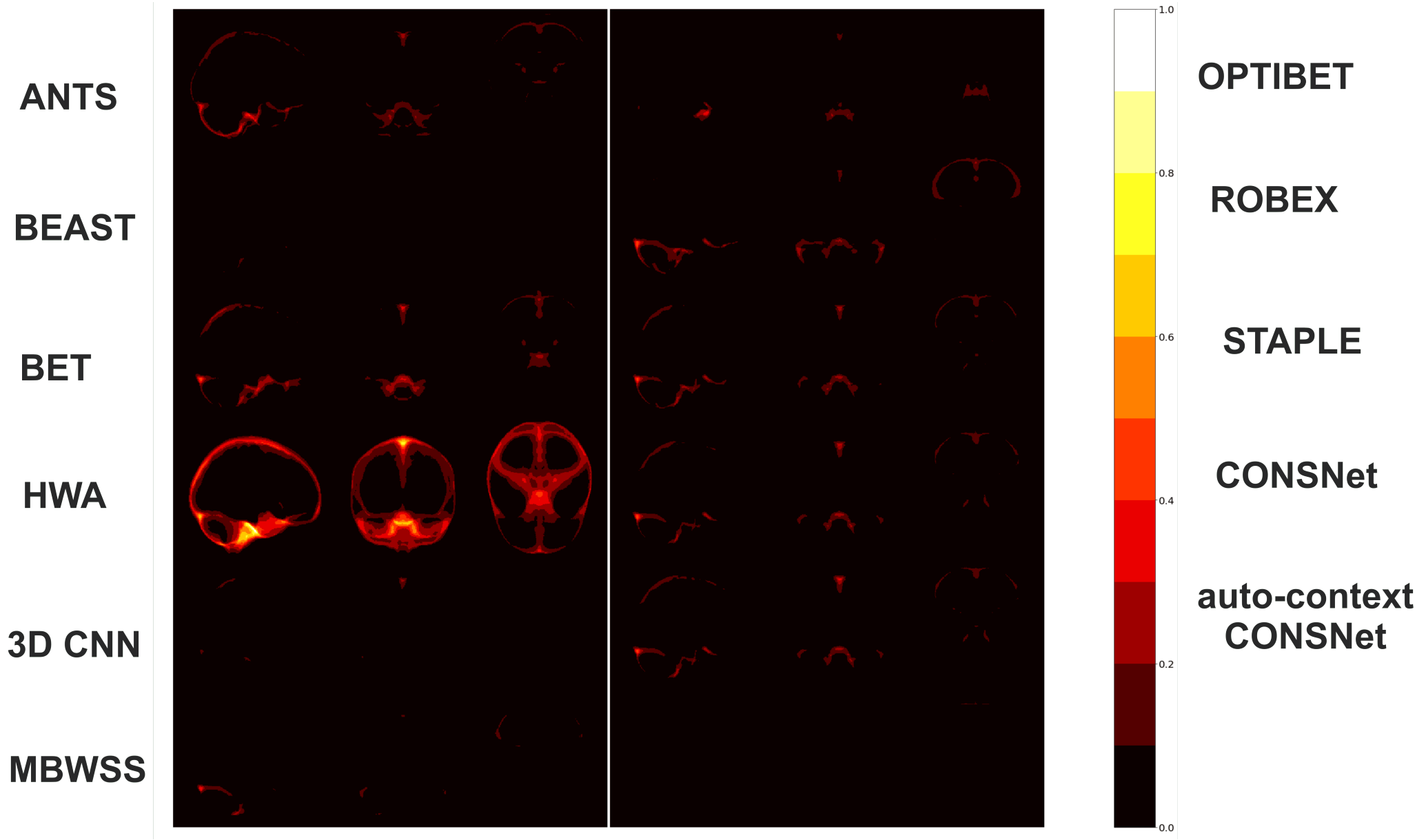

Figure 12: Sagittal, coronal and axial heat map projections for the LPBA40 dataset of FP using the manual segmentations as reference.

2-fold analysis does not prove optimal generalization. When the 2-fold models were applied to datasets which data were not included in the training step the best results regarding Dice Coefficient, for instance, were around 92%. That happens due to CNNs be a data-driven approaches and during its training stage the CNNs only learn to reproduce the annotation guidelines by the provided training data, being exclusively biased for that kind of manual annotation. As a consequence, since LPBA40 and OASIS are not robust datasets (only one vendor and scanner), the optimal generalization was not achieved.

### *4.3. General Discussion*

CONSNet is the optimal choice between our two pipelines approaches. This model is fast, low cost, and achieved similar performance compared to auto-context CONSNet in Experiment 1. Also, CONSNet had similar results to auto-net in experiment 2 but with lower cost. Comparing the results of experiment 1 and 2, our method using the silver-standard data as input has better generalization compared to the 2-fold experiment. As mentioned in the text, they are generated through an agreement among different automatic guidelines, overcoming the possible super-specialization towards one guideline in the CNN training. Also, they eliminate inter-/intra- rater variability in the annotation step. Lastly, our solution, as far as we know, is the first truly big data one for SS. We generate silver-standard masks for unlabelled data enlarging our input data and we adopted a patch-wise training having more than 100k input data.

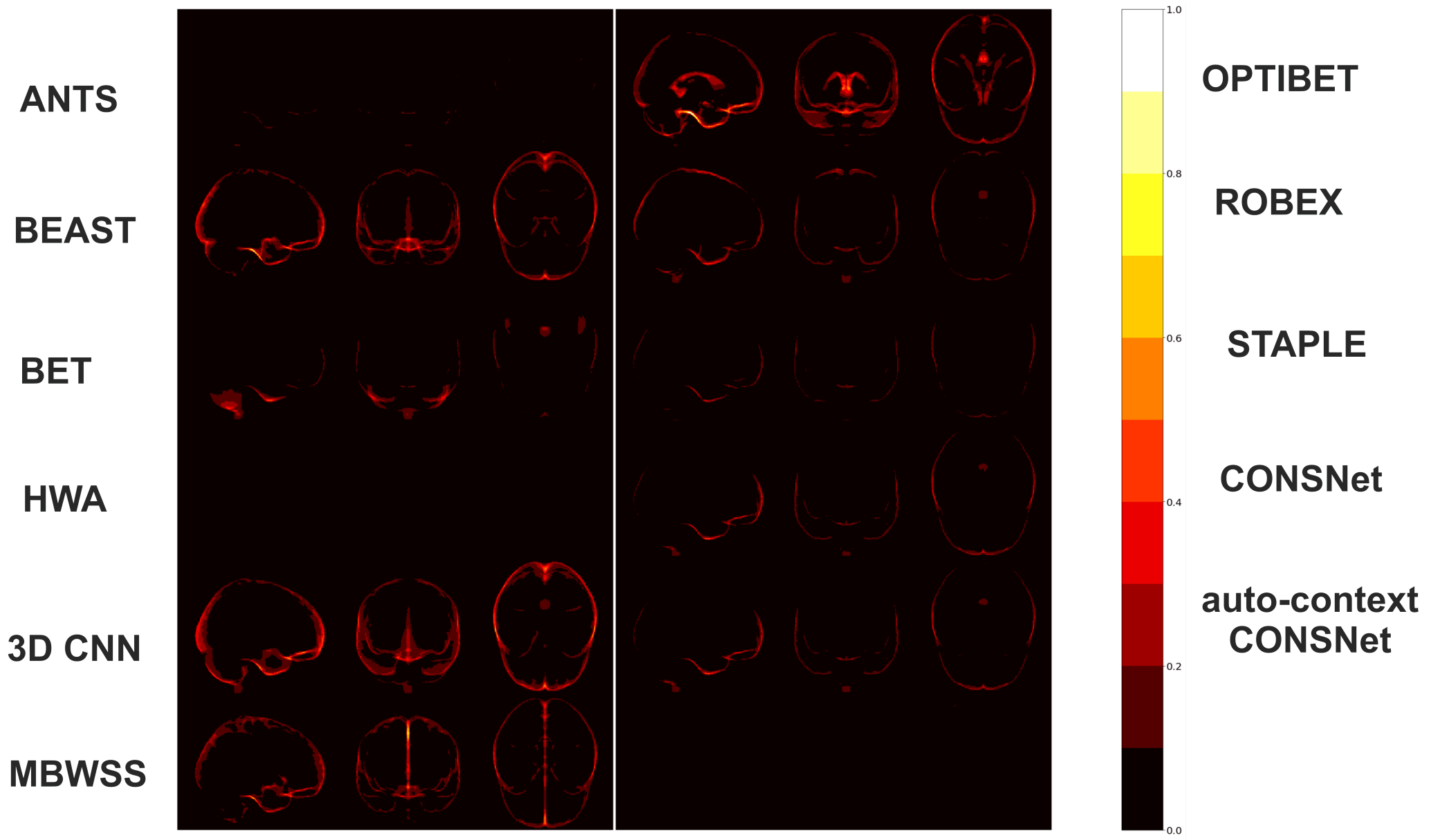

Figure 13: Sagittal, coronal and axial heat map projections for the LPBA40 dataset of FN using the manual segmentations as reference.

## 5. Conclusions

In this work, we have proposed a robust convolutional neural network for brain MR imaging SS, fully trained using silver-standards masks. The overall analysis indicated that our CONSNet and auto-context CONSNet are comparable to state-of-the-art automatic approaches, faster than the most robust non DL-based methods even under CPU processes, with an optimal generalization. Also, with the usage of silver-standard masks, we provide a low cost solution for annotated input data augmenting the training data, which reduces the inter-/intra-rater variability, and overcome the super-specialization of data-driven approaches since they are generated through a consensus agreement.

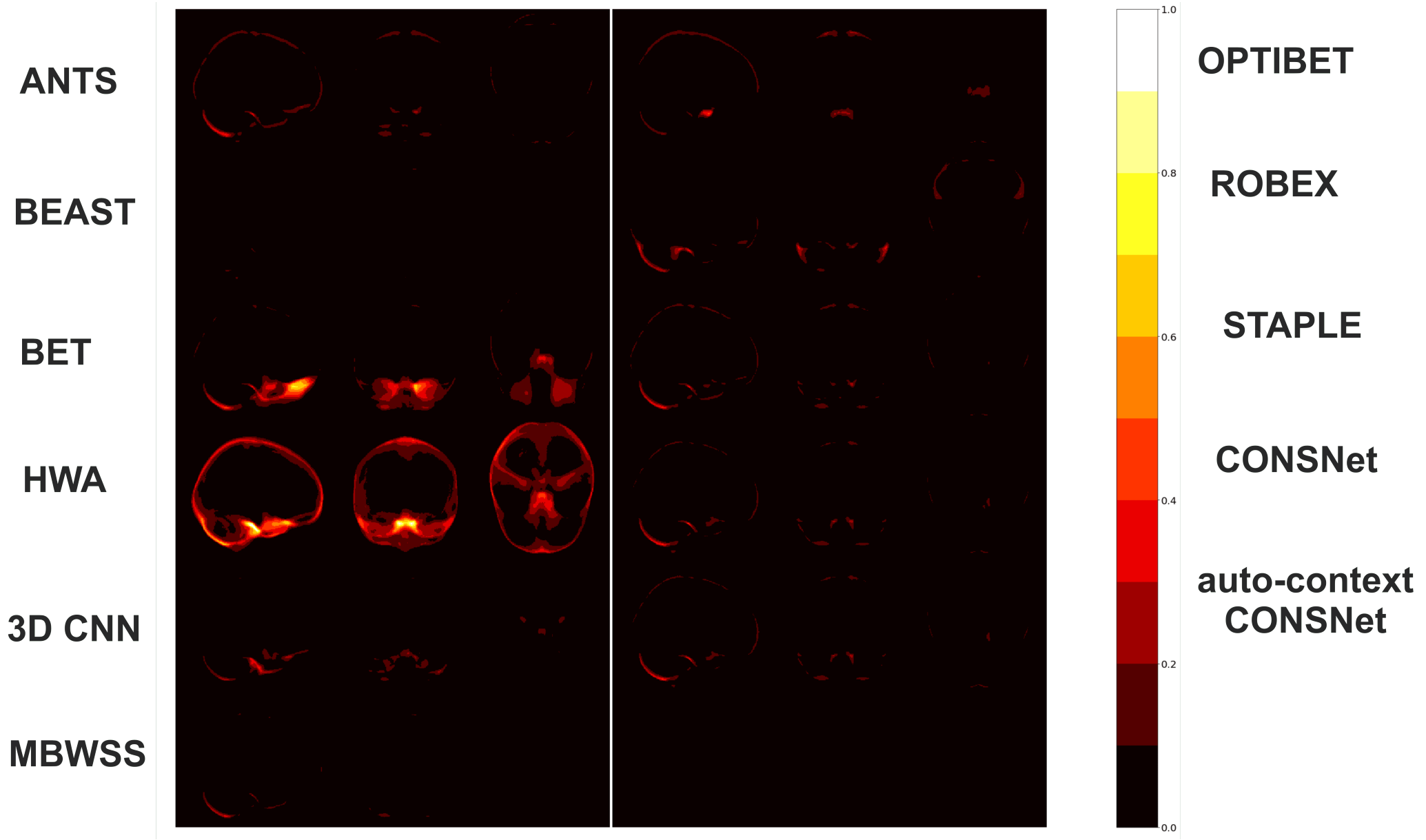

Figure 14: Sagittal, coronal and axial heat map projections for the OASIS dataset of FP using the manual segmentations as reference.

With this work we want leverage the usage of silver-standard brain masks for large-scale studies in medical image processing. We are aware the scarcity of expertly annotated data in other tasks of medical image processes, and we want in the future extend our contributions to other medical imagery applications.

## Acknowledgments

Oeslle Lucena thanks FAPESP (2016/18332-8), Roberto Souza thanks the NSERC CREATE I3T program and the Hotchkiss Brain Institute, Leticia Rittner thanks CNPq (308311/2016-7), Richard Frayne is supported by the Canadian Institutes for Health Research (CIHR, MOP-333931) and the Hopewell Professorship in Brain Imaging, and Roberto A. Lotufo thanks CNPq (311228/2014-3)

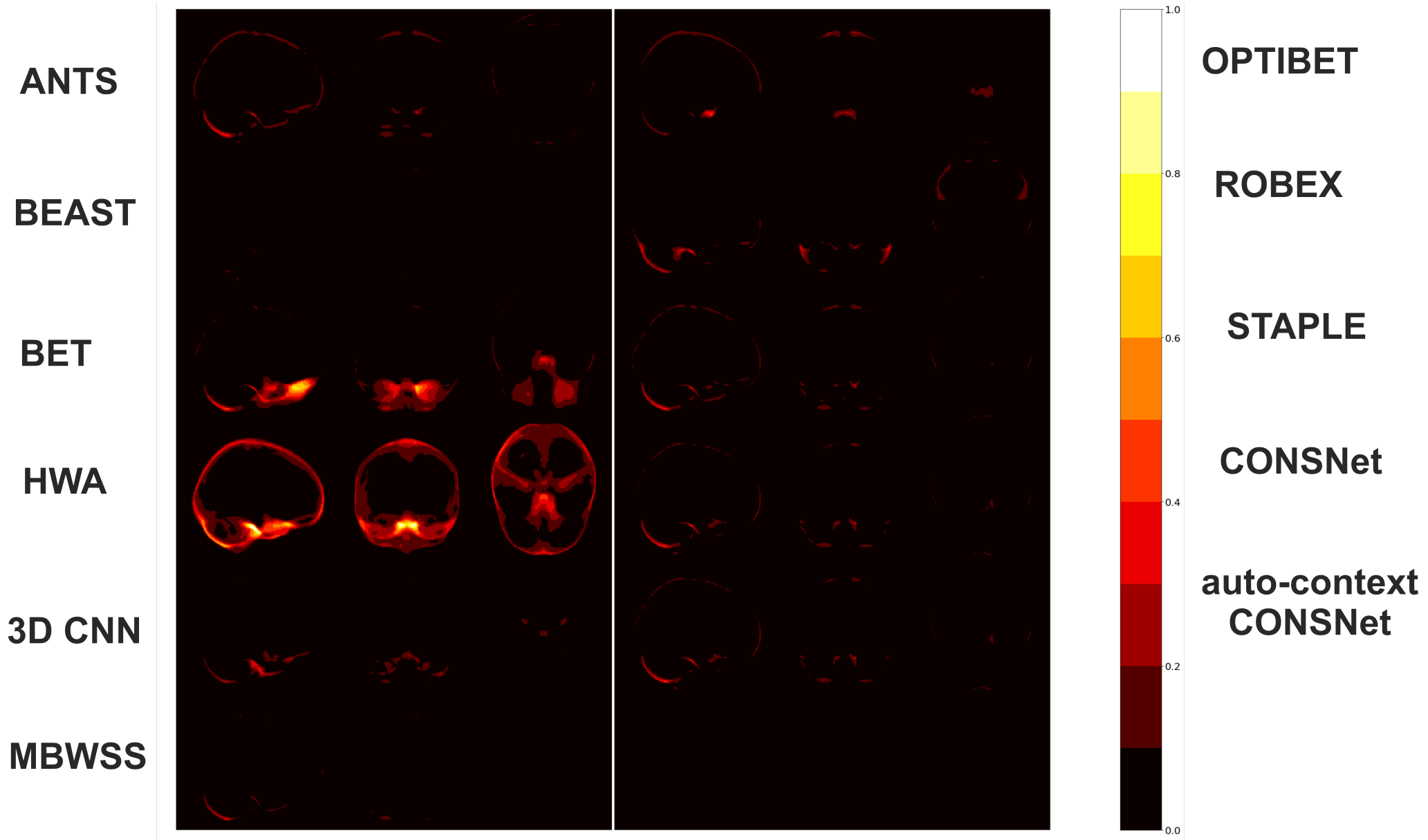

Figure 15: Sagittal, coronal and axial heat map projections for the OASIS dataset of FN using the manual segmentations as reference.

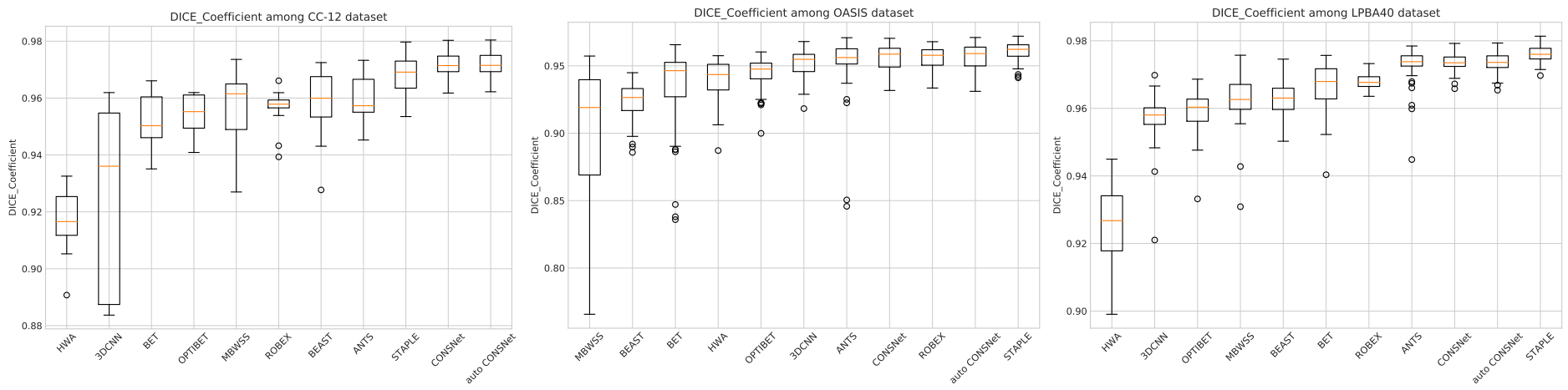

(a) (b) (c)

Figure 16: Box-plots of average Dice coefficient for each dataset analyzed (*CC-12*, OASIS, and LPBA40). BSE results were excluded due to poor results and for a better scaling of the data.

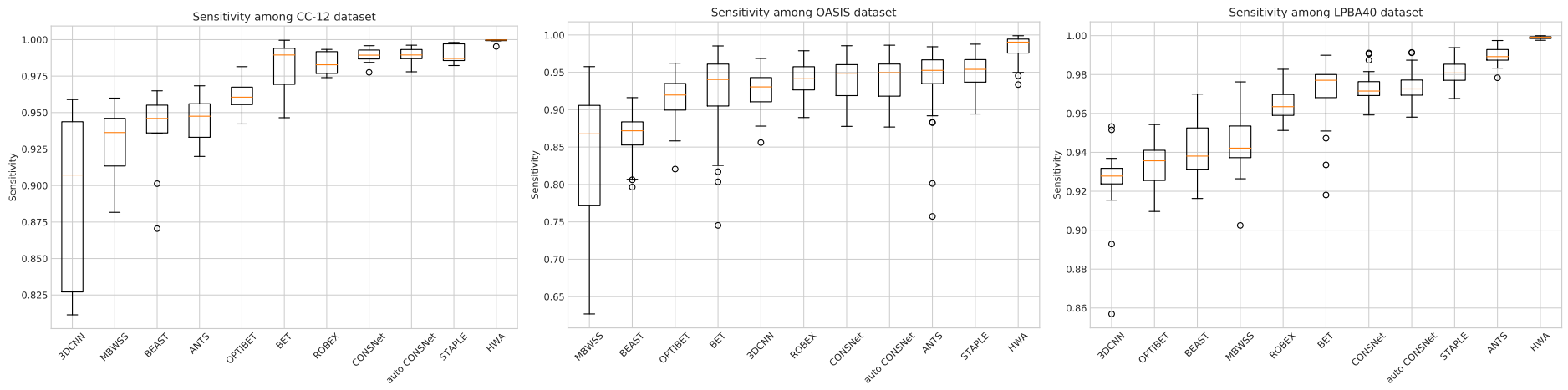

(a) (b) (c)

Figure 17: Box-plots of average Sensitivity for each dataset analyzed (*CC-12*, OASIS, and LPBA40). BSE results were excluded due to poor results and for a better scaling of the data.

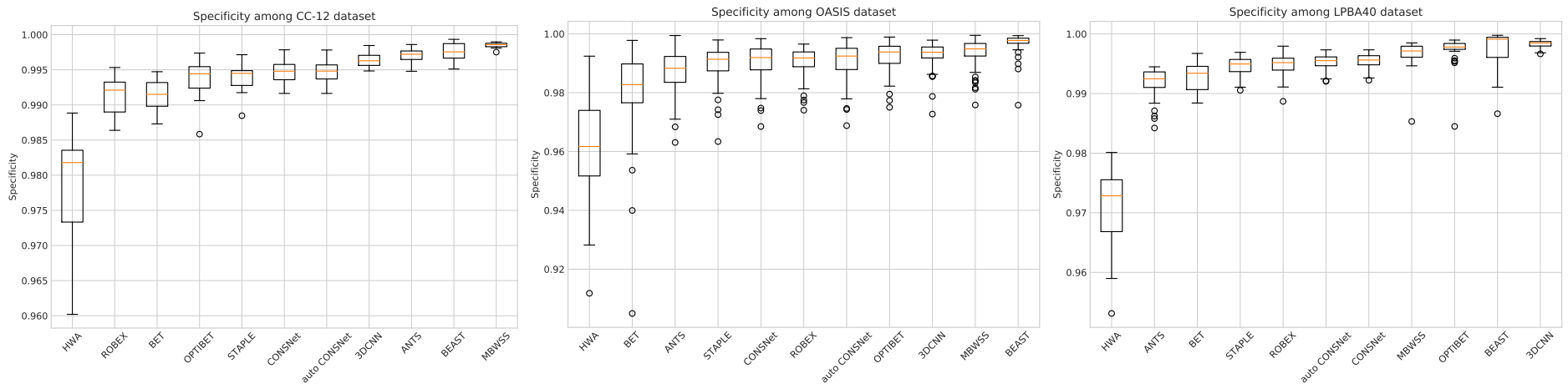

(a) (b) (c)

Figure 18: Box-plots of average Specificity for each dataset analyzed (*CC-12*, OASIS, and LPBA40). BSE results were excluded due to poor results and for a better scaling of the data.

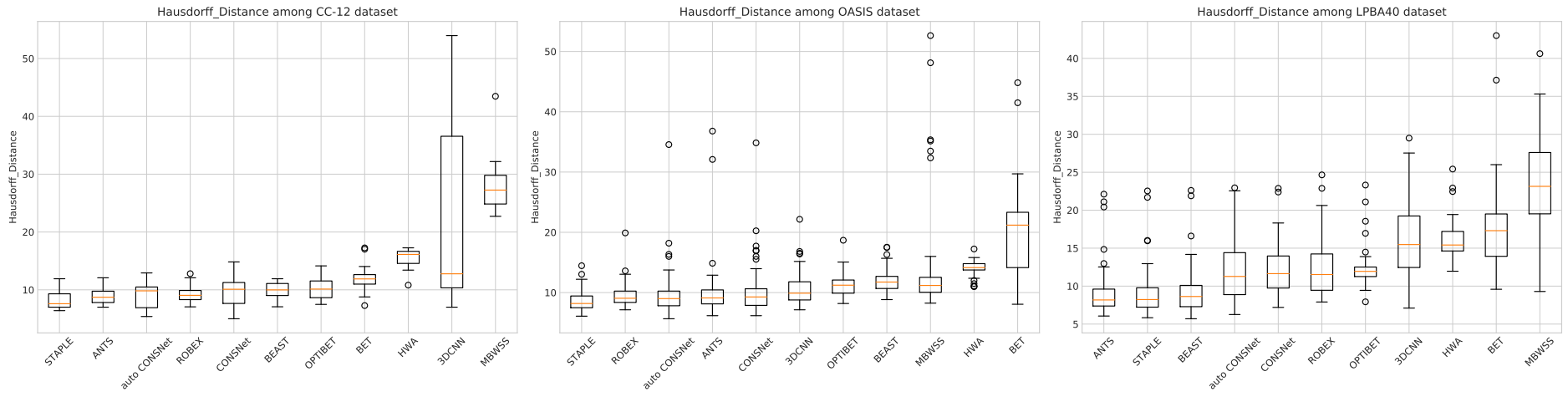

(a) (b) (c)

Figure 19: Box-plots of average Hausdorff distance for each dataset analyzed (*CC-12*, OASIS, and LPBA40). BSE results were excluded due to poor results and for a better scaling of the data.

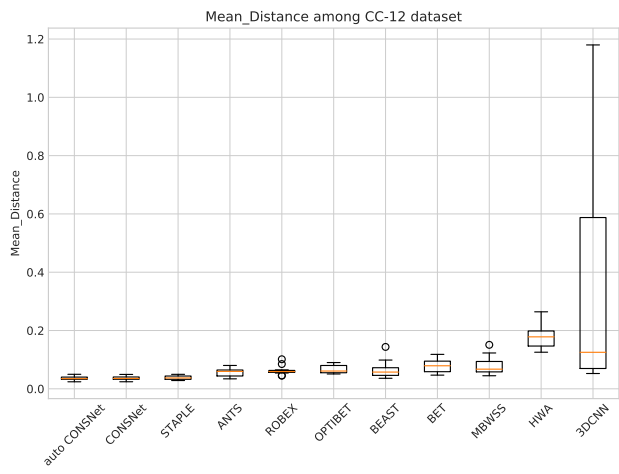

(a)

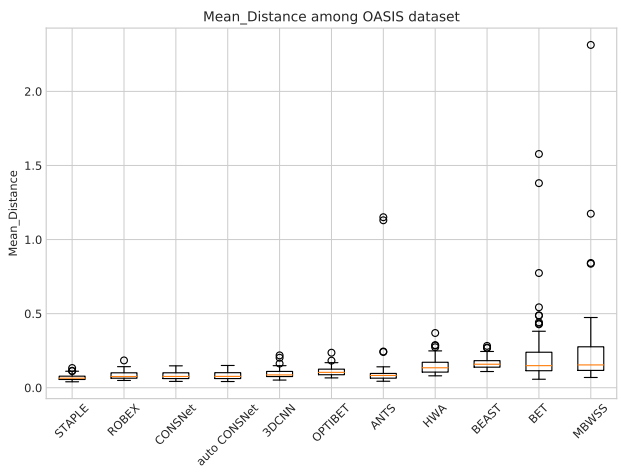

(b)

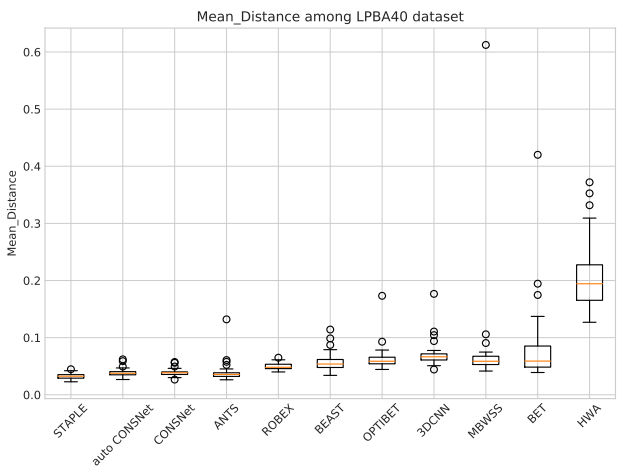

(c)

Figure 20: Box-plots of average symmetric surface-to-surface mean distance for each dataset analyzed (*CC-12*, OASIS, and LPBA40). BSE results were excluded due to poor results and for a better scaling of the data.